\documentclass{article}

 \usepackage[preprint]{neurips_2026}

\usepackage[utf8]{inputenc} 
\usepackage[T1]{fontenc}    
\usepackage{hyperref}       
\usepackage{url}            
\usepackage{booktabs}       
\usepackage{amsfonts}       
\usepackage{nicefrac}       
\usepackage{microtype}      
\usepackage{xcolor}         
\usepackage{graphicx}
\usepackage{amsmath}
\usepackage{cleveref}
\usepackage{booktabs}
\usepackage{todonotes}
\usepackage{subcaption}
\usepackage{wrapfig}

\usepackage{siunitx}
\usepackage{multirow}
\usepackage{pgfplots}

\crefname{figure}{Figure}{Figures}
\Crefname{figure}{Figure}{Figures}

\crefname{table}{Table}{Tables}
\Crefname{table}{Table}{Tables}

\crefname{section}{Section}{Sections}
\Crefname{section}{Section}{Sections}

\crefname{subsection}{Section}{Sections}
\Crefname{subsection}{Section}{Sections}

\crefname{equation}{Equation}{Equations}
\Crefname{equation}{Equation}{Equations}

\crefname{appendix}{Appendix}{Appendices}
\Crefname{appendix}{Appendix}{Appendices}

\usepackage{pifont}
\newcommand{\cmark}{\ding{51}}
\newcommand{\xmark}{\ding{55}}
\usepackage[table]{xcolor}

\title{SymTorch: Symbolic Distillation of Neural Networks}

\author{%
  Elizabeth S.Z. Tan \quad Adil Soubki \quad Miles Cranmer \\
  Department of Applied Mathematics and Theoretical Physics\\
  University of Cambridge\\
  \texttt{\{eszt2, as3591, mc2473\}@cam.ac.uk} \\
}

\begin{document}

\maketitle

\begin{abstract}
What mathematical functions do neural network components learn? Symbolic distillation addresses this question by expressing neural network components with interpretable, closed-form mathematical expressions that expose the functional structure learned during training.
We develop symbolic distillation as a systematic, architecture-agnostic methodology, and release our approach as the open-source SymTorch package -- a PySR-powered library built natively for the PyTorch ecosystem. Applying this methodology across diverse architectures, we find that SymTorch is successful in the automated discovery of physical laws. Specifically, our approach (1) recovers pairwise interaction forces from graph neural networks trained on empirical $n$-body observations, (2) distills the exact closed-form PDE/ODE solutions of multiple physical systems, including the value of constants, from physics-informed neural networks trained on sparse data, and (3) uncovers the chaotic dynamics of the Lorenz system from high-dimensional data, ultimately outperforming the base neural network on downstream prediction tasks. We further demonstrate the utility of our framework for model interpretability by providing an optimized implementation of SLIME — a symbolic extension to the LIME explainability method. SLIME consistently outperforms LIME across predictive metrics across eight popular classification and regression benchmarks, while still providing an interpretable local symbolic model. Lastly, we investigate replacing transformer MLP layers with symbolic surrogates: replacing 1--7 layers with symbolic approximations yields 2--19\% throughput improvements and up to 18.7\% VRAM reduction, with the resulting hybrid models lying on the Pareto front of throughput versus perplexity among open-source LLMs of comparable scale.
\end{abstract}

\begin{figure}[ht]
    \centering
    \includegraphics[width=1\linewidth]{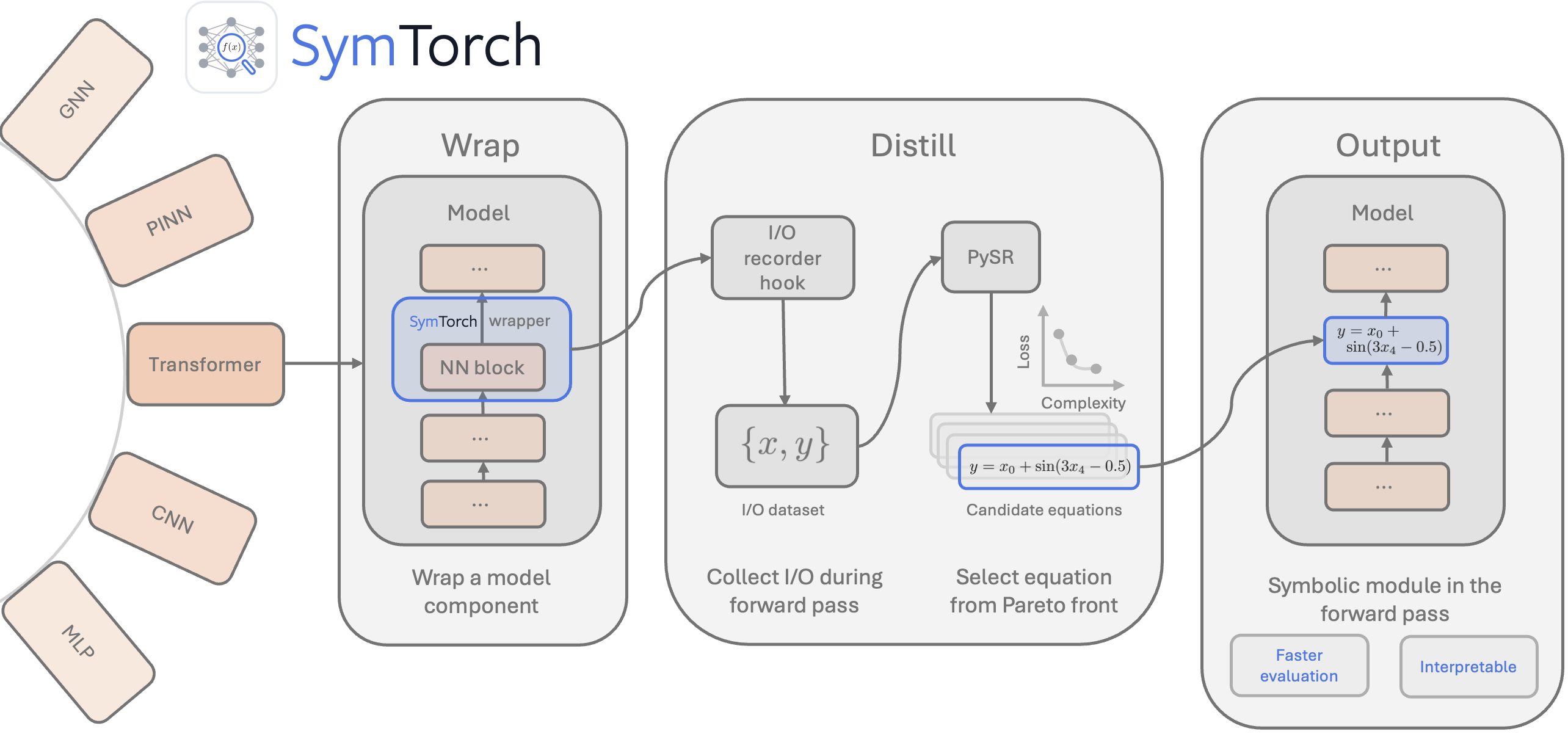}
    \caption{Overview of symbolic distillation. For a trained PyTorch model, any NN component is wrapped and its inputs and outputs (I/O) are collected during a forward pass over sample data. Symbolic regression is performed on the I/O to produce expressions approximating the component's behavior at different levels of complexity. The user can then select an equation from the Pareto front and replace the component with this expression, producing a hybrid neural-symbolic model.}
    \label{fig:conceptual_pic}
\end{figure}

\section{Introduction}
\vspace{-0.5em}
Deep learning models excel at analyzing large datasets, but remain largely uninterpretable.
In the sciences, deep models are increasingly used for physical simulation \cite{walrus, the_well, mpp, ml_fluids, aion}, where interpretability is particularly critical: the highly parameterized nature of deep networks makes it difficult to extract what physical laws the model has learned, and thus their ability to advance scientific understanding is fundamentally limited.

In physics, interpretability means using concise equations that reliably explain phenomena \cite{brunton}. These equations are written in the language of mathematical operators and variables whose physical effects are understood. Newton's laws exemplify this: $F=ma$ is accurate in idealized settings yet deliberately inexact. Adding terms for drag or relativistic effects would sacrifice interpretability for minimal accuracy gains. 

Inspired by this physics-centric view of interpretability, we bring a symbolic perspective to Neural Network (NN) analysis. We use Symbolic Regression (SR) to distill NN components into human-readable mathematical formulas. Symbolic distillation provides architecture-agnostic interpretability by approximating component behavior with closed-form expressions. This enables direct inspection of input-output mappings and analysis of how input variations affect outputs, including in out-of-distribution settings.

The user can optionally replace these components with the discovered symbolic expression directly in the forward pass, producing a \textit{live hybrid model}. Gradients can flow through the symbolic expressions and thus the model can continue to train. Crucially, we show that this hybrid model, with the neural component replaced by an equation, recovers the performance of the original network on some downstream tasks. Beyond interpretability, symbolic replacement yields practical gains in computational efficiency, reducing both inference latency and the memory requirements necessary for deploying models, as simple closed-form expressions can be faster to evaluate and take up less storage space than dense matrix operations; we explore this direction by presenting a framework to accelerate LLM inference through replacing Multi-Layer Perceptron (MLP) blocks with symbolic surrogates.

We instantiate our methodology in SymTorch, built on PyTorch and interfacing with PySR \cite{pysr}, which uses genetic algorithms to search the space of symbolic expressions. SymTorch enables systematic application of SR to deep learning models. Our work makes three main contributions:
\begin{itemize}
    \item \textbf{A general methodology for symbolic distillation of NN components} We develop a systematic approach for extracting closed-form mathematical expressions from arbitrary PyTorch modules. We release this methodology as the SymTorch package,\footnote{\texttt{pip install torch-symbolic}} the first and only open-source framework for performing per-layer symbolic distillation of PyTorch models. We further provide extensive documentation and notebooks for all experiments.\footnote{Documentation: \url{https://symtorch.readthedocs.io/en/latest/}.} By lowering the barrier to applying symbolic regression in this setting, SymTorch enables a broader set of researchers to experiment with and build upon symbolic distillation.
    \item \textbf{Experiments for data-driven scientific discovery across diverse architectures} We establish that: (1) symbolic distillation of GNN edge functions recovers known physical force laws, (2) symbolic distillation reveals the exact closed-form PDE/ODE solutions of multiple physical systems, including the heat equation, traveling wave equation, and damped harmonic oscillator, from Physics-Informed Neural Networks (PINNs) trained on sparse data, and (3) uncovers the chaotic dynamics of the Lorenz system from high-dimensional data, while outperforming the base neural network on prediction tasks.
    \item \textbf{Symbolic framework for model interpretability} SymTorch provides a symbolic extension to the LIME explainability method. By enabling the capture of nonlinear relationships, SLIME outperforms LIME across all measured predictive metrics on eight popular classification and regression benchmarks, while still providing an interpretable local symbolic model. We further show that symbolic regression of LLM-computed arithmetic reveals systematic biases corresponding to known theoretical limitations of finite-precision transformers.
    \item \textbf{Analysis of symbolic surrogates for transformer inference} We show that symbolic distillation of LLM MLP layers consistently outperforms both layer-skipping and PCA-only baselines at all operating points. Replacing 1--7 layers yields 2--19\% throughput improvements and up to 18.7\% VRAM reduction. Crucially, the perplexity costs of SR and PCA are nearly identical across most operating points, establishing that dimensionality reduction, not symbolic approximation, is the binding constraint. The resulting hybrid models lie on the Pareto front of throughput versus perplexity among similarly-sized open-source LLMs.
\end{itemize}
These results show that symbolic distillation can recover useful and interpretable mathematical structure learned by deep models across fundamentally different domains.
\vspace{-0.5em}
\section{Background \& related work}
\vspace{-0.5em}
\subsection{Interpreting model behavior}
LIME and SHAP are common model-agnostic methods of explaining whole-model behavior. SHAP \cite{shap} assigns importance scores to inputs, whereas LIME \cite{lime} fits a local linear approximation to the model by perturbing the model inputs around a point of interest.
\citet{slime} introduce Supralocal Interpretable Model-Agnostic Explanations (SLIME), a method similar to LIME. Rather than approximating local model behavior with linear models, SLIME employs symbolic surrogate models, enabling the capture of non-linear relationships. SLIME further augments the surrogate model training set with points sampled from the true data distribution, as opposed to only training on randomly perturbed samples as in LIME. SymTorch provides the first natively integrated implementation of SLIME, enabling seamless, out-of-the-box symbolic explanations for any PyTorch-based black-box model.
\subsection{Symbolic interpretability}
\vspace{-0.5em}
Symbolic distillation as an interpretability tool remains underexplored despite promising results. \citet{vafa2025foundation} showed that transformers trained on planetary dynamics, although accurate in their predictions, fail to recover Newton’s law of gravitation. Instead they exhibit sample-specific “implied” force laws, as revealed through SR of model predictions. In contrast, \citet{symbolic_dist_gnns} showed that deep networks with appropriate inductive biases, trained on empirical $n$-body data, learn physically meaningful pairwise interaction forces. These forces can be recovered by fitting a symbolic expression to the input–output mappings of model component. This procedure not only recovers known force laws but also led to the discovery of a previously unknown empirical formula for dark matter halo over-density that outperformed hand-designed models. We extend this work in \cref{subsubsec:gnns}, introducing a pruning-based approach that matches bottleneck architectures without requiring prior knowledge of system dimensionality.
\subsection{Symbolic regression and related approaches}
\vspace{-0.5em}
Several families of methods seek to discover symbolic or interpretable expressions from data. AI Feynman \cite{ai_feynman} applies physics-inspired decomposition strategies to raw observational data, while SINDy \cite{brunton, champion_2019} identifies governing equations of dynamical systems via sparse regression. SINDy requires the construction of a candidate function library, and the identified equations are constrained to be linear combinations of library terms. SR with PySR, on the other hand, makes no prior assumptions about functional form beyond the chosen operators and complexity budget.
\subsection{Symbolic Regression with PySR}
\vspace{-0.5em}
Our methodology builds on PySR, extending it for deep learning applications. PySR performs multi-population evolutionary search over analytic expressions through mutation, crossover, simplification, and constant optimization. The algorithm maintains a Pareto front of expressions that best fit the data at different complexities.
In PySR, complexity is measured as the number of nodes when $g$ is written out as an expression tree. PySR selects the best equation by balancing the complexity and accuracy of the expression. Further information on the SR algorithm and choosing the best equation is in Appendix \ref{appndx:pysr_algo}.
SymTorch provides capabilities that go beyond what PySR offers. 
\begin{itemize}
    \item \textbf{Live differentiable hybrid model}: PySR solely provides the symbolic regression engine, outputting a Pareto front of equations to approximate model behavior, but SymTorch provides a live differentiable hybrid model. This is a fundamentally different output artifact from PySR, and it is this distinction that enables the entirely new workflows that were previously not possible with just PySR.
    \item \textbf{Continued training and new workflows}: SymTorch supports continued training after symbolic module replacement. Once you swap in the symbolic model, the rest of the network can fine-tune to compensate for the errors introduced. This requires maintaining gradient flow through non-standard computational graphs. PySR cannot do this alone as it only produces a set of equations, rather than a module that can slot into a larger model for continued training.
    \item \textbf{Compute efficiency benefit}: When replacing the neural component with the symbolic surrogate model, SymTorch completely removes the original component from the computational graph, which using hooks and PySR cannot do. Hooks still require the original model to be run, so the user would not get any speedup or memory saving benefit.
\end{itemize}
\section{Framework}
We describe our methodology for symbolic distillation of neural network components: wrapping target modules, collecting input-output activations, performing SR, replacing model blocks with symbolic surrogates, and describe our implementation of SLIME for explaining local model behavior using analytical equations.
\paragraph{Distilling neural networks with SymTorch}
\texttt{SymbolicModel} is the entry point for all SymTorch functionality. Inheriting from PyTorch's \texttt{nn.Module}, it can wrap around both PyTorch NNs and callable functions, provided these functions form a mapping, \begin{wrapfigure}{r}{0.48\textwidth}
    \centering
    \includegraphics[width=1\linewidth]{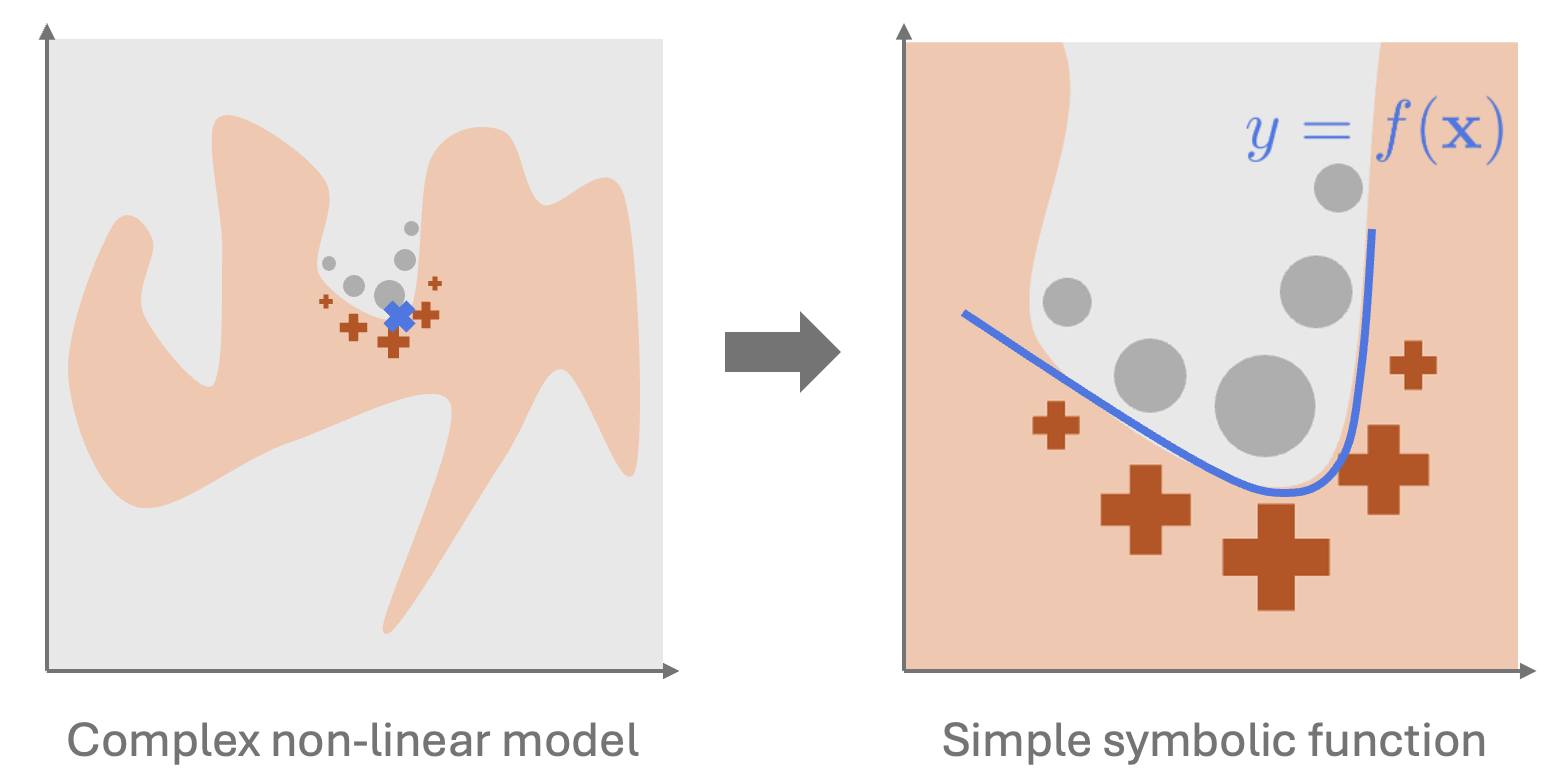}
    \caption{Approximating local model behavior with SLIME. For a complex non-linear model, we choose the point of interest $\mathbf{x}^*$. We sample points around this region and fit a symbolic model to these points.}
    \label{fig:slime_pic}
\end{wrapfigure}

\begin{equation}\label{eqn:func_mapping}
    f:\mathbf{x}_i\mapsto \mathbf{y}_i,
\end{equation}
where $\mathbf{x}_i = (x_{1,i}, \cdots ,x_{d,i})$ and $ \mathbf{y}_i = (y_{1,i}, \cdots, y_{D,i})$. SymTorch expands the SR problem in \cref{eqn:SR_problem} for many-dimensional outputs by fitting a separate symbolic model to each output dimension. We define the original function $f$ as a model \textit{block}. For PyTorch NNs, the wrapper registers forward hooks to record the block's inputs and outputs over user-supplied data.
The \texttt{distill} method performs SR by calling PySR with user-specified parameters (operators, genetic algorithm hyperparameters, fitness function). The outputs are fitted as closed-form equations of the block's input variables. 
\paragraph{Switching to symbolic approximations}
After fitting, \texttt{switch\_to\_symbolic} completely replaces the original block in the computation graph with expressions selected from the Pareto front based on their complexity, so that subsequent forward passes evaluate the symbolic equations in place of the original computation.
If no complexity is specified, SymTorch chooses the best expression that balances complexity and accuracy, according to \cref{eqn:pysr_score}. 
The model can continue training with fixed equations while other weights update. \texttt{switch\_to\_block} restores the original function.
\paragraph{SLIME implementation}
SymTorch includes a native implementation of SLIME \cite{slime} for local model explanations. Given a black-box model $f$ and a point of interest $\mathbf{x}^*$, we fit a symbolic surrogate $s \in \mathcal{S}$ to approximate $f$ in a neighborhood of $\mathbf{x}^*$, as shown in \cref{fig:slime_pic}.
The training dataset $\mathcal{D}$ consists of: (1) the $J$ nearest neighbors to $\mathbf{x}^*$ from the data distribution, and (2) $N_{\text{synthetic}}$ Gaussian-sampled points around $\mathbf{x}^*$. The Gaussian variance defaults to half the variance of the $J$ neighbors, so the approximation region is controlled by $J$. We fit $s$ by minimizing,
\begin{align*}
     s = \arg\min_{s \in \mathcal{S}} &\sum_{z_i \in \text{synthetic}} \pi_x(z_i) [f(z_i)-s(z_i)]^2 +M \sum_{z_j \in \text{neighbors}} [f(z_j)-s(z_j)]^2,
\end{align*}
where $\pi_x(z_i) = \exp(-||\mathbf{x}^*-z_i||^2/\sigma^2)$
is a proximity-weighted kernel and $M$ weights points from the true distribution. The \texttt{distill} method supports SLIME via a built-in flag. 
\section{Experiments}
In this section, we evaluate SymTorch across four settings: scientific discovery from structured architectures (\cref{subsubsec:gnns,subsubsec:chaotic_lorenz,subsubsec:pinns}), local model interpretability via SLIME (\cref{subsubsec:slime_lime}), symbolic analysis of LLM-learned arithmetic (\cref{subsec:llm_interp}), and a new framework to improve LLM performance through symbolic replacement of transformer MLP layers (\cref{sec:symdistill}).
\vspace{-0.5em}
\subsection{Recovering physical laws from GNNs}\label{subsubsec:gnns}
\vspace{-0.5em}
We apply symbolic distillation to GNNs trained on particle dynamics, recovering known physical force laws from learned message-passing representations. GNNs are well-suited to modeling $n$-body systems due to their inductive biases: they naturally represent particles as nodes and interactions as edges. We evaluate on four systems: Coulomb-like charge interactions, a spring force, and power-law repulsive forces scaling as $\propto r^{-1}$ and $\propto r^{-2}$. Building on \citet{symbolic_dist_gnns}, we introduce a dynamic pruning-based regularization that achieves comparable force-law recovery without requiring prior knowledge of the system's dimensionality. Details of the theoretical basis of this pipeline and complete experimental details including data, architecture, hyperparameters and training, and extended results are provided in Appendix \ref{appdx:gnns}
\vspace{-0.5em}
\paragraph{Symbolic distillation of GNN edge model}
For a trained GNN, we apply symbolic distillation to the edge model component, which learns the pairwise interaction forces. With our pruning-based technique, we successfully recovered the true closed-form expressions of the interaction forces across all systems.
We note that the use of a GNN is essential for this problem, as applying SR directly to the raw dataset fails to recover the true interaction forces; assume a dataset $\{(y_i,\{\mathbf{x}_{1,i}, \cdots, \mathbf{x}_{100,i}\}\}_{i=1\cdots N}$ where $y_i$ is the target variable which could for example be the acceleration of a single particle in a system of 100 particles. Assume the true relationship to be a composite function $a(\sum_{\tiny{j}} b(\mathbf{x}_i, \mathbf{x}_j))$. If the SR needs to consider $N$ equations for functions $a$ and $b$ then we must consider $N^2$ equations to get the true expression for $y$. In contrast, when $a$ and $b$ can be fitted independently -- as in our GNN architecture, where separate MLPs learn each function -- the search space decomposes, and only $2N$ candidate expressions need to be considered. Hence the GNN makes SR much more tractable \cite{symbolic_dist_gnns}.
\subsection{Uncovering chaotic dynamics from high-dimensional data}\label{subsubsec:chaotic_lorenz}
\begin{figure}[h]
    \centering
    \includegraphics[width=1\linewidth]{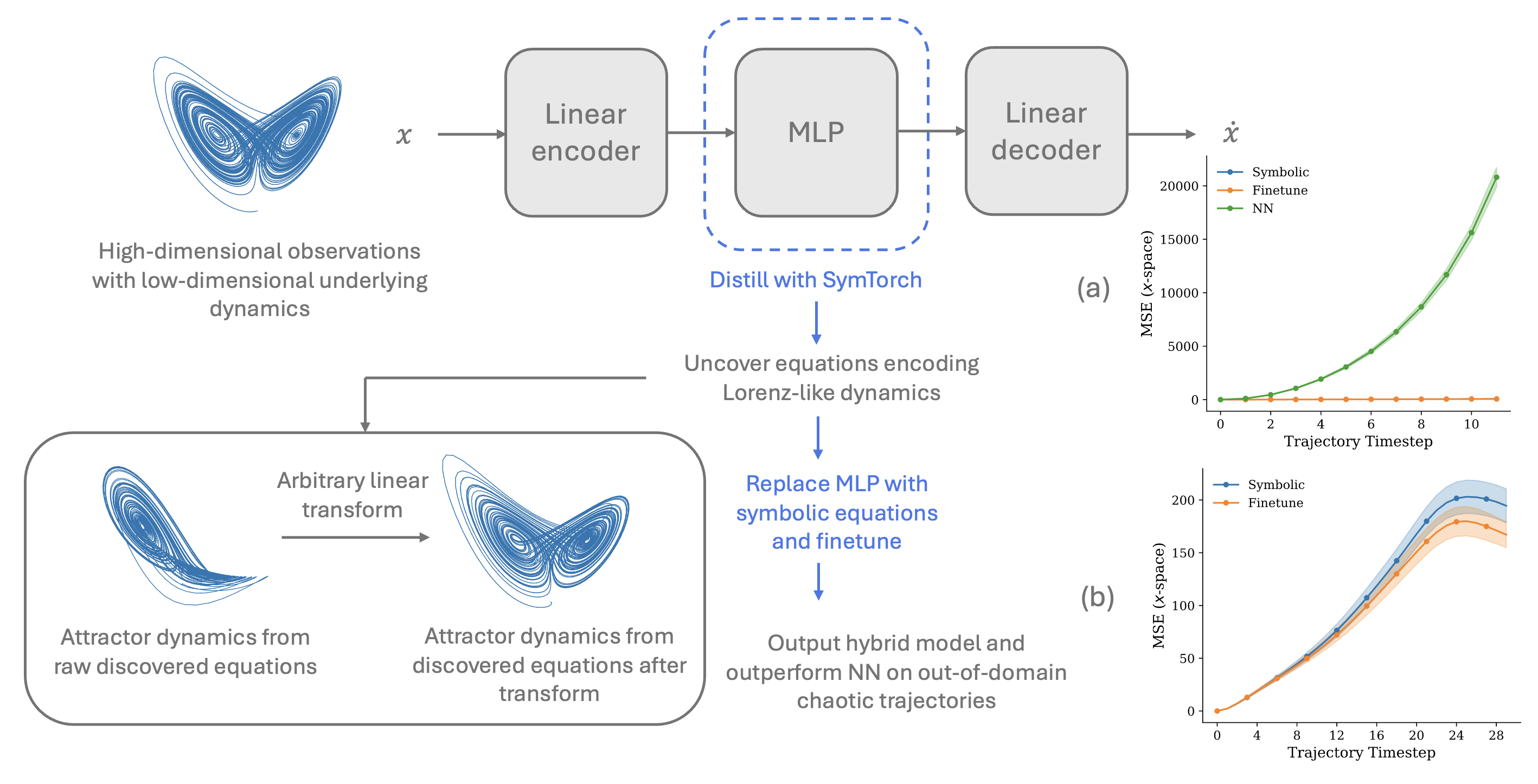}
    \caption{The model pipeline is trained to predict $\dot{x}$ from $x$, where $x$ comprises high-dimensional observations whose underlying dynamics are low-dimensional. The linear encoder projects $x$ into a latent space in which an MLP learns these simple dynamics, and a linear decoder maps the output back to produce $\dot{x}$. We apply SymTorch to distill the MLP into symbolic equations and show that the discovered equations trace out Lorenz-like attractor dynamics after an linear arbitrary transform. Replacing the MLP with these equations and fine-tuning the full model yields a hybrid that outperforms the original NN on out-of-domain chaotic trajectories, as shown in the plots on the right (MSE in $x$-space per timestep, averaged over 200 trajectories; shaded regions indicate standard error on the mean). \textbf{(b)} is the same plot as \textbf{(a)} with the NN results removed to better distinguish between the symbolic and fine-tuned model performance.}
    \label{fig:chaotic_lorenz}
\end{figure}

Symbolic regression applied to observational data assumes that parsimonious governing equations exist in the observed coordinate system — an assumption that frequently fails in practice. Experimental measurements live in high-dimensional spaces, and the intrinsic degrees of freedom of a system may be far fewer than the number of observed channels. Applying SR directly to such observations risks recovering a high-complexity expression that fits the data without reflecting any underlying sparse structure or failing to find a compact expression at all. The fundamental challenge is therefore not just to apply SR, but to do so in a coordinate system where the governing equations are genuinely sparse and interpretable. Here, we show an example of discovering low-dimensional dynamics from high-dimensional data.
\vspace{-0.5em}
\paragraph{Data pipeline, model architecture, and training}
We construct a dataset of high-dimensional observational data by projecting a single simulated Lorenz trajectory into high dimensions via a random linear map. The model architecture used is a linear encoder, followed by an MLP acting in a lower dimensional latent space, followed by a linear decoder, which is just the pseudo-inverse of the encoder. During training, the encoder learns the coordinate transform to a lower-dimensional space, while the MLP learns the Lorenz dynamics. 
\vspace{-0.5em}
\paragraph{Attractor geometry and model predictive performance}
The model successfully learns the correct attractor geometry up to an arbitrary linear transformation, though within the learned latent space the MLP can only approximate the true dynamics: the Lorenz system is chaotic, so small errors in the learned vector field compound exponentially over time. Using SymTorch, we symbolically distill the MLP into closed-form expressions; although these differ from the true Lorenz equations, a consequence of the arbitrary learned coordinate system and the MLP failing to capture the true dynamics exactly, mapping the resulting attractor back through the linear transform reproduces the characteristic two-lobe geometry. Since the linear layers were optimized with the NN in place, substituting the symbolic expressions for the MLP introduces predictions errors which can be reduced by finetuning the model with the symbolic expressions in place. All three models — neural, symbolic, and finetuned — recover the correct attractor geometry. On the single trajectory used in training the NN, the NN performs best, but averaged over 200 different trajectories with random initial conditions, both the symbolic and finetuned models achieve substantially lower prediction error (see \cref{fig:chaotic_lorenz}), suggesting the MLP overfits to single trajectories and learns piecewise heuristics that fail to generalize.
More details on data generation and model training, and extended results are in Appendix \ref{appndx:chaotic_lorenz}.

\vspace{-1em}
\subsection{Symbolic discovery of PDE solutions under data scarcity}\label{subsubsec:pinns}
We compared a PINNs with a standard NN of identical architecture for predicting (a) the temperature field governed by the 1-D heat equation, (b) the displacement field governed by the 1-D traveling wave equation, and (c) the displacement field governed by a damped harmonic oscillator. While both networks were trained directly on the same data, the PINN incorporated knowledge of the system's governing Partial Differential Equation (PDE) or Ordinary Differential Equation (ODE) for the damped harmonic oscillator, Initial Conditions (ICs), and Boundary Conditions (BCs) via additional regularization terms in its loss. Trained on only ten data points, the PINN achieved substantially better predictions than the standard network for each system as its inductive bias enforced physically consistent solutions. Using SymTorch, we further distilled the trained PINN into a closed-form analytic expression, successfully recovering the solutions to all systems tested. Using a PINN is advantageous compared to applying SR directly to the dataset, as the PINN can generate additional training data that guides SR toward equations consistent with the underlying physical solution. More information on this experiment including the precise data pipeline, model architecture, hyperparameters and training, and extended results can be found in Appendix \ref{appdx:pinns}.
\subsection{SLIME and LIME benchmarking}\label{subsubsec:slime_lime}
\begin{figure}[htbp]
    \centering
    \begin{subfigure}{0.42\textwidth}
        \centering
        \includegraphics[width=\linewidth]{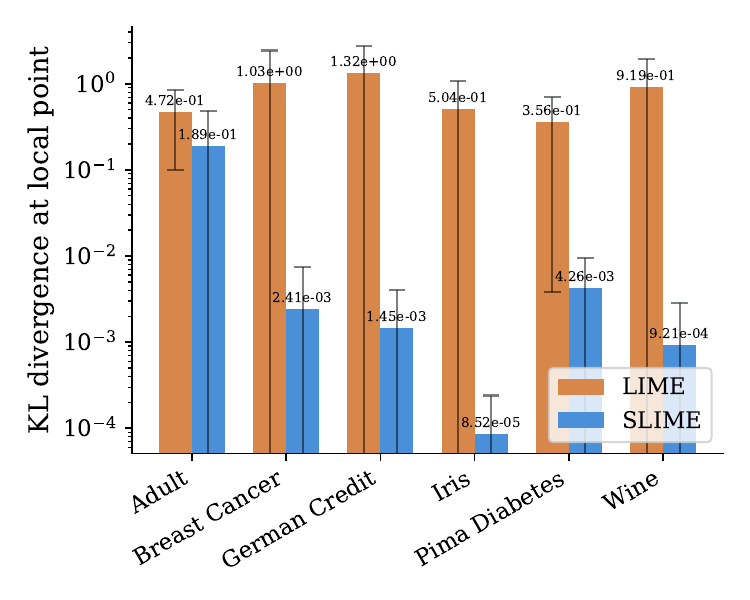}
        \caption{KL divergence between surrogate and black-box predictions across six classification datasets (log scale; lower is better).}
        \label{fig:slime_lime_class}
    \end{subfigure}
    \hfill
    \begin{subfigure}{0.42\textwidth}
        \centering
        \includegraphics[width=\linewidth]{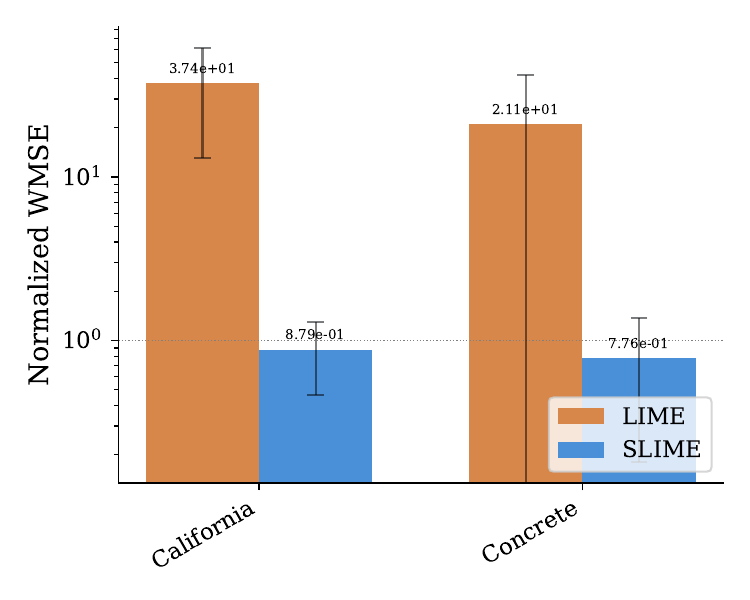}
        \caption{Normalized kernel-weighted MSE of the surrogate across two regression datasets (lower is better).}
        \label{fig:slime_lime_reg}
    \end{subfigure}
    \caption{Comparison of SLIME and LIME local surrogate fidelity across eight tabular datasets. Results are averaged over five random local points per dataset; error bars show $\pm$ 1$\sigma$.
    }
    \label{fig:slime_lime_benchmark}
\end{figure}

We benchmark SLIME against LIME on eight tabular datasets 
spanning both classification and regression tasks. For each dataset we train a black-box MLP and evaluate both surrogate methods at five independently sampled local points. At each point, LIME and SLIME each fit a surrogate model in the local neighborhood; we then compare the surrogate predictions against the black-box model's predictions on held-out points drawn from the same neighborhood.\\
For classification we report KL divergence between the surrogate's predicted class probabilities and the model's at the local point of interest. For regression we report normalized kernel-weighted MSE over the local neighborhood. \cref{fig:slime_lime_benchmark} summarizes the results. Across classification and regression benchmarks, SLIME consistently gives much higher-fidelity local approximations than LIME, due to its ability to capture more complex non-linear relationships. For more details on this benchmarking including the experimental setup, datasets, model training and hyperparameters, and extra results, see Appendix \ref{appndx:slime_lime_benchamrking}.

\vspace{-2em}
\subsection{Uncovering biases in LLM-learned arithmetic}
\label{subsec:llm_interp}
LLMs often fail at elementary numerical tasks, such that commercial systems rely on external tool calls for reliable computation \citep{dziri-etal-2023-faith, xval}. In this case study, we use SR to analyze the true function computed by the LLM. Symbolic distillation is well-suited to this task: rather than treating the LLM as a black-box that is only right or wrong, we can recover an explicit analytic approximation of the computation it is performing. This enables direct inspection of systematic numerical errors and reveals how, and where, the model's internal heuristics deviate from the intended operations.

We study the model Llama-3.2-1B-Instruct as a representative small LLM and analyze the operation the model has learned for (a) addition of two 3-digit numbers; (b) multiplication of two 3-digit numbers; (c) counting the number of 1s in a string of 1s and 0s; (d) converting Celsius to Fahrenheit. More information on the experiment including specific prompts and SR parameters can be found in \cref{appdx:llm_operations}.

The results of this analysis can be found in \cref{tab:llm_operations}. It should be noted that the correct equation was present in the Pareto front of the SR for all of these tasks except counting. However it was not the chosen 'best' equation suggesting that the LLM gets these tasks mostly correct with the addition of some systematic errors. It is known that the expressivity of finite-precision transformers \citep{chiang-etal-2023-tighter}, or even log-precision transformers \citep{merrill-sabharwal-2025-logic}, is limited to variants of first-order logic with counting or majority quantifiers. These logics cannot, in general, implement an algorithm to perform exact counting over arbitrary-length inputs, which might be related to why this task proved more difficult.
\vspace{-1em}
\begin{table}[ht]
  \caption{Symbolically distilled LLM-learned operations. The best equation, as determined by \cref{eqn:pysr_score}, is shown here. The $\epsilon$ denotes a small ($\ll 1$) term.}
  \label{tab:llm_operations}
  \centering
  \begin{small}
    \begin{tabular}{ccc}
      \toprule
      \textbf{Operation} &
      \textbf{Expected Equation} &
     \textbf{ LLM-Learned Operation} \\
      \midrule
      Addition &
      {\tiny $x_0 + x_1$} &
      {\tiny\shortstack{
      $x_1 \cdot ((\mathrm{inv}(x_0 - 70.16) + 1.07)
      +$\\ $(\mathrm{inv}(\sin(x_1) + 0.80) + x_1)
      \cdot (-\epsilon))\,x_0$
      }} \\
      \\
      Multiplication &
      {\tiny $x_0 x_1$} &
      {\tiny $x_1 (\mathrm{inv}(1.66 - 31.3\sin(x_0)) + x_0)$} \\
      \\
      Counting &
      {\tiny $x_0 + \cdots + x_5$} &
      {\tiny\shortstack{
      $(-0.11x_0 + 1.67)(x_1 +
      (x_1(x_0 +$\\$ 1.43) - 2.50)(x_2 - 0.53)
      (x_0 + x_3 - 2.15) + x_2)$
      }} \\
      \\
      Temperature Conversion &
      {\tiny $\frac{9}{5}x + 32$} &
      {\tiny
      $\mathrm{inv}(x_0 - 168.86)\mathrm{inv}(x_0 - 129.65)
      + \mathrm{inv}(x_0 - 51.87)x_0 + 33.79$
      } \\
      \bottomrule
    \end{tabular}
  \end{small}
  \vskip -0.1in
\end{table}

\subsection{Symbolic distillation for LLM performance}
\label{sec:symdistill}
\vspace{-0.5em}
A broad line of work aims to reduce overall inference cost and latency through techniques such as quantization \cite{dettmers-etal-2022-int8}, pruning \cite{lagunas2021blockpruningfastertransformers}, and speculative decoding \cite{leviathan-etal-2023-fast}. MLP layers constitute a substantial portion of transformer inference compute \citep{wei-etal-2024-building}. In this case study, we demonstrate symbolic distillation using SymTorch as a new approach for inference optimization of models.
\vspace{-0.5em}
\paragraph{Methodology}
We first apply Principal Component Analysis (PCA) to both the input and output activations of each MLP block. This allows us to produce a small set of fairly simple equations to replace the layer where the principal components taken from the input activations correspond to the number of variables provided to SR and the principal components taken from the output activations correspond to the number of equations fit in our symbolic model. Generally, it is more efficient to be harsher on reducing the dimensionality of the inputs than the outputs since SR scales worse with the number of input variables. These principal components are then used to fit the SR equations. 
We used the Qwen2.5-1.5B-Instruct model \citep{qwen2.5} for these experiments and the Wikitext-2-v1 dataset \citep{wikitext}. The dataset is split into a training, validation, and test set each consisting of 178k tokens, respectively. The training set was used to train the PCA models and provide the sample data for the SR. We used perplexity on the held out test set to quantify the changes in performance of the LLM. After performing a sensitivity analysis, we chose 32 principal components for the input reduction and 8 for the output reduction.
To select which layers to intervene on, we use a simple greedy approach. We symbolically distill all 28 layers of the network independently and rank them by their individual perplexity impact on the validation set. The symbolic surrogates are then turned on in order, starting with the lowest perplexity impact layers. Additional details, hyperparameters, results, limitations, and extensions to this study can be found in Appendix \ref{adx:symdistill}.
\begin{figure}[ht]
    \centering

    \begin{subfigure}{0.48\textwidth}
        \centering
        \includegraphics[width=\linewidth]{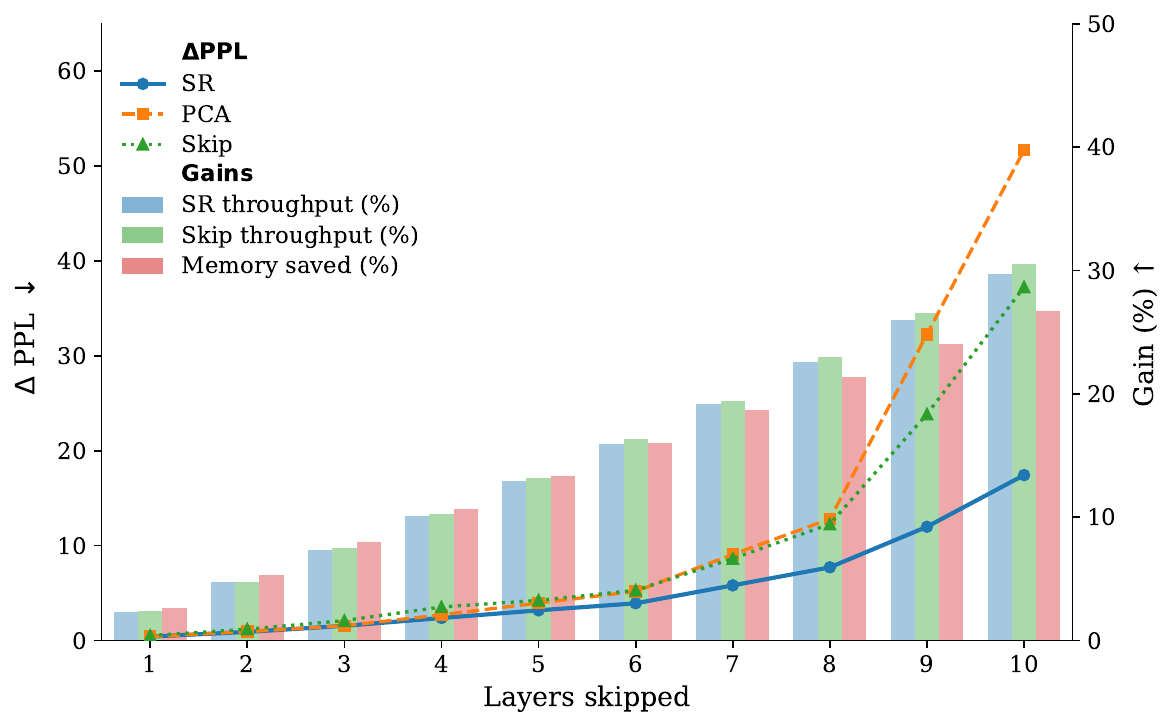}
        \caption{Perplexity degradation and efficiency gains as a function of the number of layers replaced, for SR, PCA, and skip-layer substitution methods.}
        \label{subfig:llm_distill_results}
    \end{subfigure}
    \hfill
    \begin{subfigure}{0.48\textwidth}
        \centering
        \includegraphics[width=\linewidth]{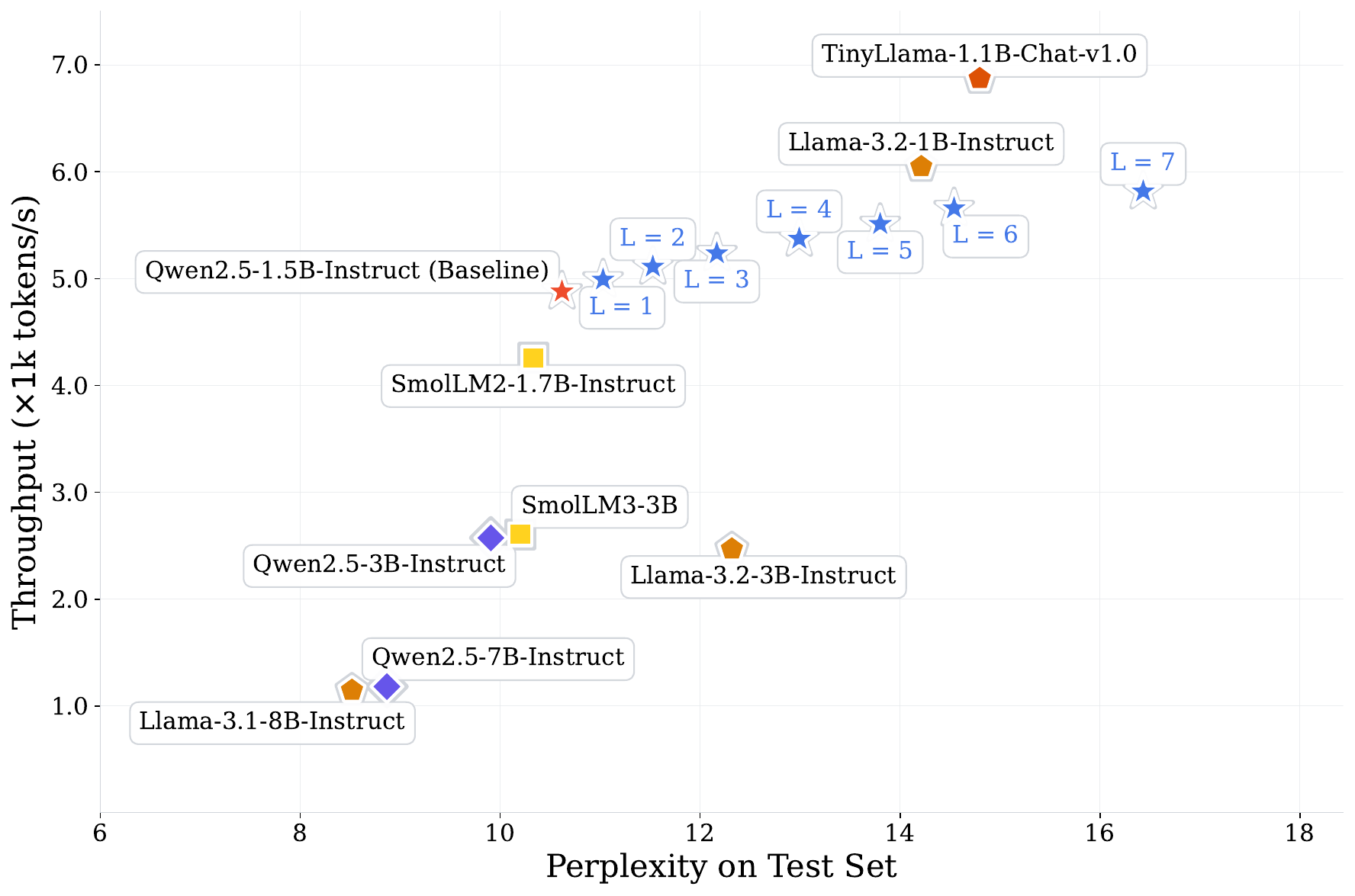}
        \caption{Throughput--perplexity trade-off across baseline LLMs and Symbolic-Hybrid variants ($L = 1$--$7$ layers replaced).}
        \label{subfig:llm_benchmarking}
    \end{subfigure}
    \caption{Symbolic-hybrid models replace the feed-forward layers of Qwen2.5-1.5B-Instruct with symbolic expressions. \textbf{(a)} shows the perplexity and efficiency trade-offs as more layers are replaced. \textbf{(b)} places the resulting models in context against standard open-source LLMs of comparable scale, demonstrating that the hybrid models with $L = 1$--$7$ layers replaced lie on the Pareto front of similarly sized open-source models. Perplexity is evaluated on the Wikitext-v2 test set.}
    \label{fig:llm_main}
\end{figure}

\vspace{-0.5em}
\paragraph{Experiments \& discussion}
We evaluate the intervention by looking at both the perplexity impact on the test set, as well as the change in model throughput (i.e., tokens/sec) and VRAM usage. For these experiments we prepare four conditions: (1) a baseline with no intervention, (2) a control where MLPs are effectively skipped, (3) a control where PCA compression is introduced at the input and output, and (4) an intervention with the symbolic surrogates replacing the MLPs. The PCA condition still runs the full dimensionality MLP to isolate the impact of the dimensionality reduction alone. We progressively skip the 10 lowest impact layers, as estimated by their perplexity impact on the validation set.
The results are shown in \cref{subfig:llm_distill_results}. Looking at the perplexity, SR clearly provides the best static intervention and improves over both the skip and PCA-only conditions. The improvement is especially strong as more layers are replaced. In terms of throughput, SR nearly matches skipping the MLP altogether. Since the interventions are trained in isolation, gains from the addition of layers are non-additive. Overall, this offers 158MB of VRAM and 2.4\% increased throughput for a 0.444 perplexity hit when replacing a single layer. If the perplexity budget allows up to 3 points (\textasciitilde10\% higher cross-entropy loss), then replacing four layers nets a 9.8\% improvement in token throughput and a 10.7\% reduction in VRAM. This could be further combined with quantization techniques. We additionally compare (see \cref{subfig:llm_benchmarking}) the resulting neuro-symbolic models, replacing up to layer 7, with several similarly-sized open weight LLMs by looking at the throughput versus perplexity trade-off. The symbolically distilled models perform competitively and lie on the Pareto frontier.
\vspace{-0.5em}
\section{Discussion}
In this section, we discuss the limitations of symbolic distillation, strategies for mitigating them, and guidance on when the approach is most appropriate. We close with our motivations for releasing this paper and SymTorch, and our hopes for its impact on future work.
\vspace{-0.5em}
\subsection{When does symbolic distillation work?}
\vspace{-0.5em}
Symbolic distillation becomes increasingly intractable as dimensionality grows, since the search space for SR expands combinatorially with the number of input variables. Pairing SR with a NN that learns a compact latent representation has shown to be an effective way of reducing the dimensionality of the distillation target.

However, dimensionality reduction alone is insufficient for scientific discovery. Without appropriate inductive biases, a neural network may distribute a single underlying function across multiple components in an entangled, piecewise fashion. Different parts of the network learn fragments of the same relationship, making it difficult to isolate and distill any one component into a physically meaningful symbolic expression. Using inductive biases to constrain components to learn distinct functions is a way to address this issue. For instance, a GNN edge model with the right architecture is forced to learn pairwise interaction forces exclusively (as in \cref{subsubsec:gnns}), and the NN module in \cref{subsubsec:chaotic_lorenz}  must learn to model the chaotic dynamics as the encoder and decoder layers do not have the flexibility to do so. Distilling such components individually then yields interpretable equations that correspond to physically meaningful quantities, rather than opaque fragments of a larger entangled computation.

For highly complex model components, SR may struggle to produce accurate global approximations. Nevertheless, it remains one of the few viable routes to functional interpretability; without it, such components would remain entirely opaque. The SLIME approach addresses this limitation: rather than seeking a single symbolic expression that captures the behavior of a component across its entire input domain, SLIME recovers simple local expressions that describe behavior in the neighborhood of a specific point. This makes symbolic approximation tractable even for components whose global behavior is too complex to compress into a closed-form expression.\
\vspace{-0.5em}
\subsection{Future work and outlook}
\vspace{-0.5em}
Symbolic distillation remains an underexplored approach to deep model interpretability. By presenting a generalized framework and releasing SymTorch as an open-source tool, we hope to accelerate progress in this area. Lowering the barrier to applying SR to a diverse range of deep learning architectures opens these techniques to a broader research community, and we look forward to seeing how others build upon them.

\section*{Acknowledgments}
The authors would like to thank Rachel C. Zhang for comments on the draft of this paper. Elizabeth S.Z. Tan's work is supported by the Aglaia Family Office. Adil Soubki thanks the UK Science and Technology Facilities Council (STFC) for a Ph.D. studentship. Miles Cranmer is grateful for support from the Schmidt Sciences AI2050 Early Career Fellowship and the Isaac Newton Trust. 

\bibliographystyle{plainnat} 
\bibliography{references}    

@misc{pysr,
      title={Interpretable Machine Learning for Science with PySR and SymbolicRegression.jl}, 
      author={Miles Cranmer},
      year={2023},
      eprint={2305.01582},
      archivePrefix={arXiv},
      primaryClass={astro-ph.IM},
      url={https://arxiv.org/abs/2305.01582}, 
}

@article{champion_2019,
   title={Data-driven discovery of coordinates and governing equations},
   volume={116},
   ISSN={1091-6490},
   url={http://dx.doi.org/10.1073/pnas.1906995116},
   DOI={10.1073/pnas.1906995116},
   number={45},
   journal={Proceedings of the National Academy of Sciences},
   publisher={Proceedings of the National Academy of Sciences},
   author={Champion, Kathleen and Lusch, Bethany and Kutz, J. Nathan and Brunton, Steven L.},
   year={2019},
   month=oct, pages={22445–22451} }

@misc{vafa2025foundation,
      title={What Has a Foundation Model Found? Using Inductive Bias to Probe for World Models},
      author={Vafa, Keyon and Chang, Peter G. and Rambachan, Ashesh and Mullainathan, Sendhil},
      year={2025},
      eprint={2507.06952},
      archivePrefix={arXiv},
      primaryClass={cs.LG},
      url={https://arxiv.org/abs/2507.06952}
}

@misc{symbolic_dist_gnns,
      title={Discovering Symbolic Models from Deep Learning with Inductive Biases}, 
      author={Miles Cranmer and Alvaro Sanchez-Gonzalez and Peter Battaglia and Rui Xu and Kyle Cranmer and David Spergel and Shirley Ho},
      year={2020},
      eprint={2006.11287},
      archivePrefix={arXiv},
      primaryClass={cs.LG},
      url={https://arxiv.org/abs/2006.11287}, 
}

@misc{battaglia,
      title={Relational inductive biases, deep learning, and graph networks}, 
      author={Peter W. Battaglia and Jessica B. Hamrick and Victor Bapst and Alvaro Sanchez-Gonzalez and Vinicius Zambaldi and Mateusz Malinowski and Andrea Tacchetti and David Raposo and Adam Santoro and Ryan Faulkner and Caglar Gulcehre and Francis Song and Andrew Ballard and Justin Gilmer and George Dahl and Ashish Vaswani and Kelsey Allen and Charles Nash and Victoria Langston and Chris Dyer and Nicolas Heess and Daan Wierstra and Pushmeet Kohli and Matt Botvinick and Oriol Vinyals and Yujia Li and Razvan Pascanu},
      year={2018},
      eprint={1806.01261},
      archivePrefix={arXiv},
      primaryClass={cs.LG},
      url={https://arxiv.org/abs/1806.01261}, 
}

@article{brunton,
author = {Kutz, J. and Brunton, Steven},
year = {2022},
month = {01},
pages = {1-17},
title = {Parsimony as the ultimate regularizer for physics-informed machine learning},
volume = {107},
journal = {Nonlinear Dynamics},
doi = {10.1007/s11071-021-07118-3}
}

@misc{dist_gnns_code_repo,
    title = {Discovering Symbolic Models from Deep Learning with Inductive Biases Code Repository},
    author={Miles D. Cranmer},
    url = {https://github.com/MilesCranmer/symbolic_deep_learning}, 
year = {2020}
}

@misc{pytorch_geo,
      title={Fast Graph Representation Learning with PyTorch Geometric}, 
      author={Matthias Fey and Jan Eric Lenssen},
      year={2019},
      eprint={1903.02428},
      archivePrefix={arXiv},
      primaryClass={cs.LG},
      url={https://arxiv.org/abs/1903.02428}, 
}

@misc{adam,
      title={Adam: A Method for Stochastic Optimization}, 
      author={Diederik P. Kingma and Jimmy Ba},
      year={2017},
      eprint={1412.6980},
      archivePrefix={arXiv},
      primaryClass={cs.LG},
      url={https://arxiv.org/abs/1412.6980}, 
}

@misc{lime,
      title={"Why Should I Trust You?": Explaining the Predictions of Any Classifier}, 
      author={Marco Tulio Ribeiro and Sameer Singh and Carlos Guestrin},
      year={2016},
      eprint={1602.04938},
      archivePrefix={arXiv},
      primaryClass={cs.LG},
      url={https://arxiv.org/abs/1602.04938}, 
}

@misc{shap,
      title={A Unified Approach to Interpreting Model Predictions}, 
      author={Scott Lundberg and Su-In Lee},
      year={2017},
      eprint={1705.07874},
      archivePrefix={arXiv},
      primaryClass={cs.AI},
      url={https://arxiv.org/abs/1705.07874}, 
}

@inproceedings{slime,
author = {Fong, Kei Sen and Motani, Mehul},
title = {SLIME: Supralocal Interpretable Model-Agnostic Explanations via Evolved Equation-Based Surrogates},
year = {2025},
isbn = {9798400714641},
publisher = {Association for Computing Machinery},
address = {New York, NY, USA},
url = {https://doi.org/10.1145/3712255.3726721},
doi = {10.1145/3712255.3726721},
booktitle = {Proceedings of the Genetic and Evolutionary Computation Conference Companion},
pages = {267–270},
numpages = {4},
location = {NH Malaga Hotel, Malaga, Spain},
series = {GECCO '25 Companion}
}

@misc{wikitext,
      title={Pointer Sentinel Mixture Models},
      author={Stephen Merity and Caiming Xiong and James Bradbury and Richard Socher},
      year={2016},
      eprint={1609.07843},
      archivePrefix={arXiv},
      primaryClass={cs.CL}
}

@misc{qwen2.5,
    title = {Qwen2.5: A Party of Foundation Models},
    url = {https://qwenlm.github.io/blog/qwen2.5/},
    author = {Qwen},
    month = {September},
    year = {2024}
}

@misc{walrus,
      title={Walrus: A Cross-Domain Foundation Model for Continuum Dynamics}, 
      author={Michael McCabe and Payel Mukhopadhyay and Tanya Marwah and Bruno Regaldo-Saint Blancard and Francois Rozet and Cristiana Diaconu and Lucas Meyer and Kaze W. K. Wong and Hadi Sotoudeh and Alberto Bietti and Irina Espejo and Rio Fear and Siavash Golkar and Tom Hehir and Keiya Hirashima and Geraud Krawezik and Francois Lanusse and Rudy Morel and Ruben Ohana and Liam Parker and Mariel Pettee and Jeff Shen and Kyunghyun Cho and Miles Cranmer and Shirley Ho},
      year={2025},
      eprint={2511.15684},
      archivePrefix={arXiv},
      primaryClass={cs.LG},
      url={https://arxiv.org/abs/2511.15684}, 
}

@misc{the_well,
      title={The Well: a Large-Scale Collection of Diverse Physics Simulations for Machine Learning}, 
      author={Ruben Ohana and Michael McCabe and Lucas Meyer and Rudy Morel and Fruzsina J. Agocs and Miguel Beneitez and Marsha Berger and Blakesley Burkhart and Keaton Burns and Stuart B. Dalziel and Drummond B. Fielding and Daniel Fortunato and Jared A. Goldberg and Keiya Hirashima and Yan-Fei Jiang and Rich R. Kerswell and Suryanarayana Maddu and Jonah Miller and Payel Mukhopadhyay and Stefan S. Nixon and Jeff Shen and Romain Watteaux and Bruno Régaldo-Saint Blancard and François Rozet and Liam H. Parker and Miles Cranmer and Shirley Ho},
      year={2025},
      eprint={2412.00568},
      archivePrefix={arXiv},
      primaryClass={cs.LG},
      url={https://arxiv.org/abs/2412.00568}, 
}

@misc{mpp,
      title={Multiple Physics Pretraining for Physical Surrogate Models}, 
      author={Michael McCabe and Bruno Régaldo-Saint Blancard and Liam Holden Parker and Ruben Ohana and Miles Cranmer and Alberto Bietti and Michael Eickenberg and Siavash Golkar and Geraud Krawezik and Francois Lanusse and Mariel Pettee and Tiberiu Tesileanu and Kyunghyun Cho and Shirley Ho},
      year={2024},
      eprint={2310.02994},
      archivePrefix={arXiv},
      primaryClass={cs.LG},
      url={https://arxiv.org/abs/2310.02994}, 
}

@misc{ml_fluids,
      title={Machine learning in fluid dynamics: A critical assessment}, 
      author={Kunihiko Taira and Georgios Rigas and Kai Fukami},
      year={2025},
      eprint={2508.13430},
      archivePrefix={arXiv},
      primaryClass={physics.flu-dyn},
      url={https://arxiv.org/abs/2508.13430}, 
}

@misc{aion,
      title={Semantic search for 100M+ galaxy images using AI-generated captions}, 
      author={Nolan Koblischke and Liam Parker and Francois Lanusse and Irina Espejo Morales and Jo Bovy and Shirley Ho},
      year={2025},
      eprint={2512.11982},
      archivePrefix={arXiv},
      primaryClass={astro-ph.IM},
      url={https://arxiv.org/abs/2512.11982}, 
}

@misc{dziri-etal-2023-faith,
      title={Faith and Fate: Limits of Transformers on Compositionality}, 
      author={Nouha Dziri and Ximing Lu and Melanie Sclar and Xiang Lorraine Li and Liwei Jiang and Bill Yuchen Lin and Peter West and Chandra Bhagavatula and Ronan Le Bras and Jena D. Hwang and Soumya Sanyal and Sean Welleck and Xiang Ren and Allyson Ettinger and Zaid Harchaoui and Yejin Choi},
      year={2023},
      eprint={2305.18654},
      archivePrefix={arXiv},
      primaryClass={cs.CL},
      url={https://arxiv.org/abs/2305.18654}, 
}

@misc{xval,
      title={xVal: A Continuous Numerical Tokenization for Scientific Language Models}, 
      author={Siavash Golkar and Mariel Pettee and Michael Eickenberg and Alberto Bietti and Miles Cranmer and Geraud Krawezik and Francois Lanusse and Michael McCabe and Ruben Ohana and Liam Parker and Bruno Régaldo-Saint Blancard and Tiberiu Tesileanu and Kyunghyun Cho and Shirley Ho},
      year={2024},
      eprint={2310.02989},
      archivePrefix={arXiv},
      primaryClass={stat.ML},
      url={https://arxiv.org/abs/2310.02989}, 
}

@inproceedings{merrill-sabharwal-2025-logic,
 author = {Merrill, William and Sabharwal, Ashish},
 booktitle = {Advances in Neural Information Processing Systems},
 editor = {A. Oh and T. Naumann and A. Globerson and K. Saenko and M. Hardt and S. Levine},
 pages = {52453--52463},
 publisher = {Curran Associates, Inc.},
 title = {A Logic for Expressing Log-Precision Transformers},
 url = {https://proceedings.neurips.cc/paper_files/paper/2023/file/a48e5877c7bf86a51395.ab23b360498-Paper-Conference.pdf},
 volume = {36},
 year = {2023}
}

@misc{chiang-etal-2023-tighter,
      title={Tighter Bounds on the Expressivity of Transformer Encoders}, 
      author={David Chiang and Peter Cholak and Anand Pillay},
      year={2023},
      eprint={2301.10743},
      archivePrefix={arXiv},
      primaryClass={cs.LG},
      url={https://arxiv.org/abs/2301.10743}, 
}

@misc{wei-etal-2024-building,
      title={Building on Efficient Foundations: Effectively Training LLMs with Structured Feedforward Layers}, 
      author={Xiuying Wei and Skander Moalla and Razvan Pascanu and Caglar Gulcehre},
      year={2024},
      eprint={2406.16450},
      archivePrefix={arXiv},
      primaryClass={cs.CL},
      url={https://arxiv.org/abs/2406.16450}, 
}

@misc{leviathan-etal-2023-fast,
      title={Fast Inference from Transformers via Speculative Decoding}, 
      author={Yaniv Leviathan and Matan Kalman and Yossi Matias},
      year={2023},
      eprint={2211.17192},
      archivePrefix={arXiv},
      primaryClass={cs.LG},
      url={https://arxiv.org/abs/2211.17192}, 
}

@misc{dettmers-etal-2022-int8,
      title={LLM.int8(): 8-bit Matrix Multiplication for Transformers at Scale}, 
      author={Tim Dettmers and Mike Lewis and Younes Belkada and Luke Zettlemoyer},
      year={2022},
      eprint={2208.07339},
      archivePrefix={arXiv},
      primaryClass={cs.LG},
      url={https://arxiv.org/abs/2208.07339}, 
}

@misc{lagunas2021blockpruningfastertransformers,
      title={Block Pruning For Faster Transformers}, 
      author={François Lagunas and Ella Charlaix and Victor Sanh and Alexander M. Rush},
      year={2021},
      eprint={2109.04838},
      archivePrefix={arXiv},
      primaryClass={cs.LG},
      url={https://arxiv.org/abs/2109.04838}, 
}

@article{scikit-learn,
  title={Scikit-learn: Machine Learning in {P}ython},
  author={Pedregosa, F. and Varoquaux, G. and Gramfort, A. and Michel, V.
          and Thirion, B. and Grisel, O. and Blondel, M. and Prettenhofer, P.
          and Weiss, R. and Dubourg, V. and Vanderplas, J. and Passos, A. and
          Cournapeau, D. and Brucher, M. and Perrot, M. and Duchesnay, E.},
  journal={Journal of Machine Learning Research},
  volume={12},
  pages={2825--2830},
  year={2011}
}

@inproceedings{ai_feynman,
      title={{AI Feynman}: A physics-inspired method for symbolic regression},
      author={Silviu-Marian Udrescu and Max Tegmark},
      booktitle={Science Advances},
      volume={6},
      number={16},
      pages={eaay2631},
      year={2020},
      doi={10.1126/sciadv.aay2631},
}

@misc{moseley_pinns,
    author = {Ben Moseley},
    title = {Harmonic Oscillator PINNs}, 
    year = {2021},
    howpublished = {GitHub},
    url = {https://github.com/benmoseley/harmonic-oscillator-pinn}
}

\appendix

\section{Symbolic Regression with PySR detail}\label{appndx:pysr_algo}
\subsection{Algorithm overview}
PySR implements the following genetic algorithm (summarized from \cite{pysr}): 
\begin{enumerate}
    \item We begin with several independent populations of individual equations. This allows expressions in each population to evolve simultaneously.
    \item In each population, a tournament selection process is run: a random subset of individuals (usually two) are evaluated on their 'fitness', a metric combining both expression accuracy and complexity. The complexity of an expression is determined by the number of nodes on the binary expression tree.
    \item The fittest individual in the subset is chosen as the winner with probability $p$; otherwise the next fittest is chosen with the same probability, and so on. 
    \item A copy of the winning individual undergoes some form of mutation, where a node on the expression tree can be changed, a node added or removed. Or the individual may undergo a cross-over operation with the next fittest expression, where parts of the expression tree may swap between these two individuals. This mutation may be accepted or rejected with a probability relating to the fitness of the mutated expression.
    \item A set number of tournaments constitutes a round of evolution for the population. At the end of each evolution round, expressions may be simplified to an equivalent form (for example $x+x+x \rightarrow3*x$) or the constants may be optimized. 
    \item After a specified number of evolutions, individuals may migrate between populations.
\end{enumerate}
At each iteration, the Pareto front is updated with the best-performing equations at different complexity levels.
\subsection{Problem formulation}
For some metric $\mathcal{L}: \mathbb{R}^N \times \mathbb{R}^N\rightarrow \mathbb{R}$, inputs $\mathbf{x}_i = (x_{1,i}, \cdots ,x_{d,i}) \in \mathcal{X}\subseteq \mathbb{R}^{d}$ and target variables $y_i \subset \mathcal{Y} \in \mathbb{R}$ for $i=1,\cdots N$, SR aims to find $g$ 
\begin{equation}\label{eqn:SR_problem}
    g = \arg\min_{g \in \mathcal{S}} \sum^N_i \mathcal{L} (y_i, g(\mathbf{x}_i)),
\end{equation}
where $S$ is the set of closed-form analytic expressions. Metric $\mathcal{L}$ is the data-fitting loss (eg. mean-squared error) that often times contains a penalty on the complexity of $g$. 
\subsection{Choosing the best equation}
PySR selects the best equation by maximizing the fractional drop in log mean absolute
error relative to an increase in model complexity. Specifically, it chooses expression $j$ along on the Pareto front that maximizes the score given by
\begin{equation}\label{eqn:pysr_score}
    \text{score}_j = -\frac{\log (\text{loss}_j/\text{loss}_{j-1})}{\text{complexity}_j-\text{complexity}_{j-1}}.
\end{equation}

\section{SymTorch extra details}\label{appndx:symtorch_details}
The source code for SymTorch is available at \url{https://github.com/astroautomata/SymTorch}.
\subsection{Default PySR parameters}
Users configure SR via the \texttt{sr\_params} dictionary in \texttt{distill} (defaults in \cref{tab:def_sr_params}). These parameters go directly to PySR's \texttt{PySRRegressor}. Parameters for the \texttt{fit} method (e.g., variable names or complexities) use the \texttt{fit\_params} dictionary.
\begin{table}[ht]
  \centering
  \caption{Default SymTorch SR configurations.}\label{tab:def_sr_params}
  \vspace{0.5\baselineskip}
  \begin{tabular}{lc}
    \toprule
    \textbf{SR Parameter} & \textbf{Configuration} \\
    \midrule
    Binary operators        & \texttt{+, *} \\
    Unary operators         & \texttt{inv(x)=1/x, sin, exp} \\
    Extra sympy mappings    & \texttt{"inv": lambda x: 1/x} \\
    Number of iterations    & 400 \\
    Complexity of operators & \texttt{sin: 3, exp: 3} \\
    \bottomrule
  \end{tabular}
\end{table}

\subsection{Saving SR Results and symbolic models}
SymTorch saves PySR SR results in \texttt{SR\_output/mlp\_name}, organized by output dimension and timestamp. SymTorch works seamlessly with PyTorch's \texttt{torch.save} and \texttt{torch.load} functions.

\section{Experimental details and extended results for \cref{subsubsec:gnns}}\label{appdx:gnns}
The code repository for this experiment is located at \url{https://github.com/astroautomata/SymTorch_symbolic_distillation_GNNs}.
\subsection{Setup}
\paragraph{Model Architecture}
We follow the notation as outlined in \citet{battaglia}. There are two MLPs involved in the GNN. The first is the edge model (or edge function), $\phi^e:\mathcal{V}\times\mathcal{V}\rightarrow\mathcal{E}$, where $\mathcal{V}\subset \mathbb{R}^{L^v}$ and $ \mathcal{E} \subseteq \mathbb{R}^{L^e}$. ${L^v}$ and ${L^e}$ are the number of node features and the dimensionality of the messages respectively. In a forward pass of the GNN, the edge model takes the node features, $\mathbf{v} \in \mathcal{V}$, of two connected nodes as inputs and outputs the edge message, $\mathbf{e}_k' \in \mathcal{E}$, where $k$ denotes the edge. We concatenate the node features when inputting them into the MLP. This function can be written as 
\begin{equation}\label{eqn:edge_model}
    \mathbf{e}_k' = \phi^e(\mathbf{v}_{r_k}, \mathbf{v}_{s_k}),
\end{equation}
\noindent where $\mathbf{v}_{r_k}$ denotes the features of the receiving node, and $\mathbf{v}_{s_k}$ denotes the sending node of edge $k$. The messages to receiving node $i$ are aggregated through an element-wise summation,
\begin{equation}
    \mathbf{\bar{e}}_i'=\sum_{j\neq i} \phi^e(\mathbf{v}_i, \mathbf{v}_j),
\end{equation}
where $\mathbf{\bar{e}}_i'$ is the aggregated message.

The second MLP involved in the GNN is the node model (or node function), $\phi^v: \mathcal{V}\times\mathcal{E}\rightarrow \mathcal{D}$, where $\mathcal{D}\subseteq \mathbb{R}^D$ and $D$ is the dimensionality of the target variable, the updated node features. This model outputs the updated node features for a specific node. It takes the node features and the aggregated message for that node as an input and calculates the node update, $\mathbf{\hat{v}}_i'$.
\begin{equation}
    \mathbf{\hat{v}}_i'=\phi^v(\mathbf{v}_i, \mathbf{\bar{e}}_i'),
\end{equation} \label{eqn:node_model}
\vspace{-1em}
\paragraph{Setup}
Node features encode the state of each particle as
\begin{equation}\label{eqn:input_data}
    \mathbf{v}_i = [x_i,\, y_i,\, \dot{x}_i,\, \dot{y}_i,\, q_i,\, m_i],
\end{equation}
where $(x_i, y_i)$ are the 2D position, $(\dot{x}_i, \dot{y}_i)$ are the discretized velocities (displacement per simulation timestep $\Delta t$), $q_i$ is the charge, and $m_i$ is the mass. The model takes these features as input and predicts the particle accelerations $\mathbf{\hat{v}}_i'$, so the output dimensionality matches that of the system (2D). We evaluate the framework on four pairwise interaction forces: (a) gravitational, (b) spring, (c) $1/r$, and (d) $1/r^2$, where $r$ is the inter-particle displacement. Training minimizes the mean absolute error between predicted and true accelerations.

The training data were generated using the official code repository of \citet{dist_gnns_code_repo}. Each sample is a tensor of state information for four interacting particles; the corresponding targets are particle accelerations, computed as the negative gradient of the pairwise potential divided by mass. Each dataset comprised 10,000 simulations of 1,000 timesteps, subsampled every fifth step to reduce temporal correlation, yielding one million samples in total. These were split into training, validation, and test sets in a 70/15/15 ratio.
\subsection{Derivation: Edge Messages as Linear Transformations of the True Forces}\label{appdx:gnn_derivation}
For a GNN that has learned to predict accelerations from particle properties, the message vectors are linear transformations of the true forces, provided that the message dimension is equal to the dimensionality of the forces. The mathematical derivation for this is, as first presented in \cite{symbolic_dist_gnns}, is shown below.

In Newtonian mechanics, the resultant force, $\bar{\mathbf{f}}_i$, acting on particle $i$ is equal to the sum of the individual forces, $\mathbf{f}_k$:
\begin{equation}
    \sum_k \mathbf{f}_k = \bar{\mathbf{f}}_i.
\end{equation}
If we ignore the particle mass, the node model predicts the resultant force on the receiver particle $r_k$:
\begin{equation}\label{eqn:force_sum}
    \bar{\mathbf{f}}_i=\mathbf{\hat{v}}_i'=\phi^v(\mathbf{v}_i, \mathbf{\bar{e}}_i')=\phi^v \left(\mathbf{v}_i, \sum_{r_k=i}\mathbf{e}_k' \right).
\end{equation}
If we only consider a single interaction, and hence a single edge, the force is:
\begin{equation}\label{eqn:force_msg_single_interaction}
    \bar{\mathbf{f}}_i=\mathbf{f}_{k,r_k=i}= \phi^v (\mathbf{v}_i, \mathbf{e}_{k,r_k=i}').
\end{equation}
Again, the resultant force is the sum of the individual forces, so we can use the above equation in the many-particle case and equate this to \cref{eqn:force_sum}. Explicitly,
\begin{equation}
    \sum_{r_k=i} \phi^v (\mathbf{v}_i, \mathbf{e}_{k}')=\phi^v \left(\mathbf{v}_i, \sum_{r_k=i}\mathbf{e}_k' \right)=\bar{\mathbf{f}}_i,
\end{equation}
which demonstrates that $\phi^v$ is a linear function in its second argument:
\begin{equation}
    \phi^v(\mathbf{v}_i, \mathbf{e}_{a}'+\mathbf{e}_{b}')=\phi^v(\mathbf{v}_i, \mathbf{e}_{a}')+\phi^v(\mathbf{v}_i,\mathbf{e}_{b}').
\end{equation}
Provided $\phi^v$ is invertible in $\mathbf{e}_{k}'$, which is true when the dimensionality of $\mathbf{e}_{k}'$ matches the dimensionality of $\mathbf{\hat{v}}_i'$, then you can invert \cref{eqn:force_msg_single_interaction}:
\begin{equation}
    \mathbf{e}_{k}'=(\phi^v(\mathbf{v}_i,\cdot))^{-1}(\mathbf{f}_k).
\end{equation}
Hence, the messages $\mathbf{e}_{k}'$ are just linear transformations of the true forces $\mathbf{f}_k$. If we constrain the message dimensionality to match that of the physical system, we can fit the learned messages using a linear regression on the true forces. A strong linear correspondence indicates that the model has successfully captured the underlying physical forces.

\subsection{Model Variants}
Details of the GNN variants trained in this case study are as follows:
\begin{enumerate}
    \item \textbf{Standard}. Consisted of a GNN where the message dimension, $L^e$, was set to 100.
    \item \textbf{Bottleneck}. The message dimension was set to match the dimensionality of the system, which was 2 in all the experiments.
    \item \textbf{L1}. We applied L1 regularization, $\mathcal{L}_e$ to the edge messages,
    \begin{equation}
    \mathcal{L}_{L1} = \frac{\alpha_{L1}}{N^e} \sum_{k=0}^{N^e-1}|\mathbf{e}_k'|,
    \end{equation}
    where $N^e$ is the total number of parameters in the edge messages, and $\alpha_{L1}=10^{-2}$ is the regularization constant. The dimension of the message vectors were set to be 100, the same as the standard model variation. This regularization encourages sparse representation of messages by the absolute values of the message components, effectively driving many of them toward zero.
    \item \textbf{Kullback–Leibler (KL)}. We added a standard Gaussian prior, $\mathcal{N} \sim (\mathbf{0},\mathbf{1})$, to the components of the messages. Unlike the other model variants, the edge model in this case outputs both the mean and log-variance for each message element, which doubles the output dimensionality. This allows the messages to be treated as samples from a Gaussian distribution rather than fixed feature vectors:
    \begin{equation}\label{eqn:variational_msg}
        \mathbf{e}_k' \sim \mathcal{N}(\boldsymbol{\mu}'_k, \textbf{diag}[{\boldsymbol{\sigma}'}_k^{2}]),
    \end{equation}
    where $\boldsymbol{\mu}'_k =\phi_{\mu}^e$ and ${\boldsymbol{\sigma}'}_k^{2} = \exp(\phi_{\sigma^2}^e)$. We trained the model such that $\phi_\mu^e$ are all even outputs and $\phi_{\sigma^2}^e$ are all odd outputs from the edge model. During training, the edge messages are sampled from the distribution defined by \cref{eqn:variational_msg} before being aggregated and inputted to the node model. A regularization term equivalent to the KL-divergence between the standard normal prior and the probability distribution defined in \cref{eqn:variational_msg} is added to the loss:
    \begin{equation}\label{eqn:KL_loss_term}
        \mathcal{L}_{KL} =  \frac{1}{N^e} \sum_{k=0}^{N^e-1} \sum_{j=0}^{L^e-1} \frac{1}{2}({\mu'}_{k,j}^2 + {\sigma'}_{k,j}^2-\log({\sigma'}_{k,j}^2)).
    \end{equation}
    The KL regularization term encourages sparsity in the messages by penalizing deviations from a standard normal distribution, effectively pushing the learned mean and variance of each message component toward zero mean and unit variance. If the model is in an evaluation setting, we do not sample and just take the message elements to equal the means.
    \item \textbf{Pruning}. The dimensionality of the edge messages is gradually reduced during training until it matches that of the system using SymTorch's pruning functionality. To prune the MLP, we chose a random 10,240 data points from the validation set as sample data points and inputted these through the MLP. The dimensions with the lowest variance across these sample points were deemed as 'unimportant' and we zero-masked these dimensions. We used a Cosine Annealing pruning schedule, with pruning completing at 65\% of the way through training. SymTorch contains its own built-in method to prune MLPs in this way. This model variation is an extension to the original paper. Pruning is a similar model variation to bottleneck, but has the advantage that we do not require knowledge of the dimensionality of the system beforehand.

\end{enumerate}
\subsection{Training}
\paragraph{Model Configuration}
To create and train the GNNs, we used PyTorch and PyTorch Geometric \cite{pytorch_geo}. In all experiments, the edge model and node model MLPs contained three hidden layers each with 300 hidden units and ReLU activations between each layer. 
\paragraph{Data Augmentation}
The node model outputs predictions of the instantaneous accelerations of particles, as shown in \cref{eqn:node_model}. Before passing the node features into the model, we augment the data by adding Gaussian noise with a standard deviation of 3 to the position coordinates of all nodes simultaneously. This follows the approach used in the original paper's code and was likely introduced to improve model robustness by reducing overfitting to precise spatial positions, while also simulating the presence of noise in real-world data.
\paragraph{Predictions and Loss}
The base loss on our model was calculated to be the mean absolute error between the predicted accelerations, $\hat{\mathbf{v}}_i'$, and the actual accelerations, $\mathbf{v}_i'$, from our dataset:
\begin{equation}\label{eqn:base_loss}
    \mathcal{L}=\frac{1}{N^v}\sum^{N^v-1}_{i=0}|\mathbf{v}_i'-\hat{\mathbf{v}}_i'|.
\end{equation}
L2 regularization was also used in every training instance,
\begin{equation}
    \mathcal{L}_{L2}= \frac{\alpha_{L2}}{N^l} \sum_{l=0}^{N^l}|w_l|^2,
\end{equation}
where $N^l$ are the total number of network parameters, denoted $w_l$, and $\alpha_{L2} = 10^{-8}$ is the L2 regularization constant. Hence, the total loss is $\mathcal{L}+\mathcal{L}_{L2}$ for the standard, bottleneck and pruning model variant. For the L1 model variation, the loss is $\mathcal{L}+\mathcal{L}_{L2} +\mathcal{L}_{L1}$, and for the KL model variation the loss is $\mathcal{L}+\mathcal{L}_{L2} +\mathcal{L}_{KL}$. 
\paragraph{Training}
To train our models, we performed gradient descent using the Adam \cite{adam} optimizer with a Cosine Annealing learning rate scheduler. Each model was trained for 100 epochs with a batch size of 64 on a training set of 700,000 samples, resulting in approximately 1.1 million optimization steps. The training and validation loss was monitored every epoch.
\subsection{Symbolic distillation with SymTorch}
We used SymTorch to perform the SR on the edge model. For the standard and L1 model variant, we performed SR on the top two most important dimensions as determined by SymTorch's \texttt{get\_importance} method. Whereas for the KL model variation, we chose the two most importance message dimensions as the ones with the highest KL divergence as calculated in \cref{eqn:KL_loss_term}.\\
\paragraph{Variable transforms}
To improve the efficiency of the SR, we performed the following variable transforms on the input data: $\Delta x = x_1-x_2$, $\Delta y = y_1-y_2$, and $r=\sqrt{\Delta x^2+\Delta y^2}+10^{-2}$. We added a small constant to the distance $r$ to match the form used in the original potential equations when generating the dataset. Thus the inputs to the SR were these transformed variables as well as $m_1,m_2,q_1,q_2$.\\
\paragraph{SR parameters}
The operators that were allowed in the SR were $+,-,\times, \texttt{inv}(\cdot)$, which had complexity of 1, as well as \texttt{exp} and \texttt{log}, which we set to have complexity of 3. The complexity of constants and input variables were set to be 1. We chose a random set of 5,000 examples from the test set for the SR. For all of the tests, we ran the SR for 7,000 iterations. The parsimony argument was set to 0.05 and the maximum size of equations permitted (measured in terms of complexity) was set to 25. Lastly, we also set constraints of \texttt{exp} and \texttt{log} to be one.

\subsection{Results}
\begin{figure}[h!]
    \centering
    \includegraphics[width=0.5\linewidth]{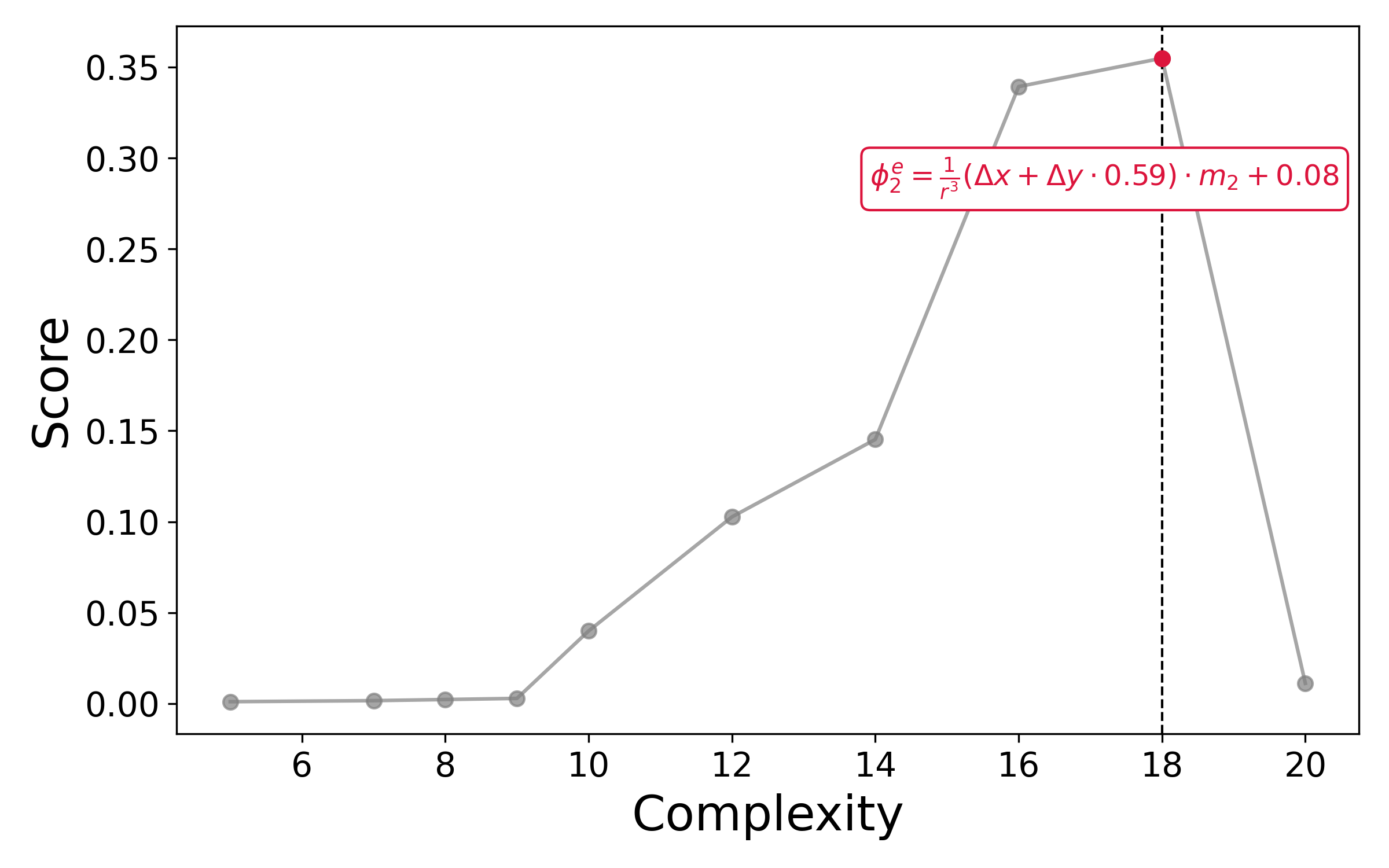}
    \caption{The score of the equations found from PySR for the pruning model trained on the $r^{-2}$ as given in \cref{tab:pareto_front}. The score is calculated from \cref{eqn:pysr_score}. The best equation, highlighted in red, produces the largest drop in $\log(\text{loss})$ per unit of additional complexity.}
    \label{fig:complexity_curve}
\end{figure}

The final loss on the trained GNNs are shown in \cref{tab:test_set_results} and the symbolic regression results are shown in \cref{tab:def_sr_params}. An example Pareto front of equations is shown in \cref{tab:pareto_front} and the corresponding scores are shown in \cref{fig:complexity_curve}.
\begin{table}[ht]
    \centering
    \caption{Test set mean absolute error (MAE) for different GNN model variants across four force law simulations. Lower values indicate better performance. The pruning variant (introduced by SymTorch) achieves comparable performance to the bottleneck model while automatically discovering the optimal  dimension. Variance values provide scale context for the prediction errors.}
    \vspace{0.5\baselineskip}
    \begin{tabular}{ccccccc}
        \toprule
        \textbf{Simulation} & \textbf{Standard} & \textbf{Bottleneck} & \textbf{L1} & \textbf{KL} & \textbf{Pruning} & \textit{\textbf{Variance}}\\
        \midrule
        Charge & 18.20 & 19.12 & 18.06 & 39.80 & 19.40 & 56417.66 \\
        $r^{-1}$     & 0.26  & 0.25  & 0.32  & 15.40 & 0.28 & 87.69\\
        $r^{-2}$     & 24.07 & 25.25 & 21.67 & 57.80 & 23.77 & 98641.90\\
        Spring & 0.24  & 0.16  & 0.20  & 7.29  & 0.23 & 55.84\\
        \bottomrule
    \end{tabular}
    \label{tab:test_set_results}
\end{table}

\begin{table}[ht]
\centering
\caption[Symbolic regression results for GNN message components.]{
Symbolic regression results for each message component.
\cmark = correct form of force law recovered; \xmark = failure.
$^{*}$ Correct form, but with a small constant added to a term (e.g., $1/(r+\text{const.})$).
$^{\dagger}$ Correct form, but only $\Delta y$ apparent in both messages.
$^{\ddagger}$ Correct form with $\Delta x$ in one message and $\Delta y$ in the other.
$^{\mathsection}$ Correct form with only $\Delta x$ or $\Delta y$ in at least one message.
}
\vspace{0.5\baselineskip}
\begin{tabular}{ccccccc}
\toprule
\textbf{Simulation} & \textbf{Message} & \textbf{Standard} & \textbf{Bottleneck} & \textbf{L1} & \textbf{KL} & \textbf{Pruning} \\
\midrule
Charge & 1 & \xmark & \cmark$^{\ddagger}$ & \xmark & \xmark & \cmark \\
       & 2 & \xmark & \cmark$^{\ddagger}$ & \cmark$^{\mathsection}$ & \xmark & \cmark$^{\mathsection}$ \\
\midrule
$r^{-1}$ & 1 & \xmark & \cmark & \cmark & \xmark & \cmark \\
         & 2 & \cmark & \cmark & \cmark & \xmark & \cmark$^{*}$ \\
\midrule
$r^{-2}$ & 1 & \cmark$^{\mathsection}$ & \cmark & \cmark$^{\mathsection}$ & \cmark$^{\ddagger}$ & \cmark \\
         & 2 & \xmark & \cmark & \xmark & \cmark$^{\ddagger}$ & \cmark \\
\midrule
Spring & 1 & \cmark$^{\dagger}$ & \cmark & \cmark & \cmark$^{\ddagger}$ & \cmark \\
       & 2 & \cmark$^{\dagger}$ & \cmark & \cmark$^{\mathsection}$ & \cmark$^{\ddagger}$ & \cmark \\
\bottomrule
\end{tabular}
\label{tab:SR_results}
\end{table}

\paragraph{Examples of successful reconstructions} Below are some examples of successful reconstructions of the true forces:
\begin{itemize}
    \item Spring; bottleneck
    \begin{equation*}
        \texttt{msg1} = \left(\frac{\texttt{1}}{\texttt{r}}-\texttt{0.99950946} \right) \cdot ( \texttt{0.8855752} \Delta \texttt{y} + \texttt{1.8560125} \Delta \texttt{x})+ \texttt{0.031805687}
    \end{equation*}
    \item $r^{-2}$; L1
    \begin{equation*}
        \texttt{msg1} = \texttt{m}_\texttt{2}((\Delta \texttt{y}+\texttt{0.43513915})+\Delta \texttt{x})\cdot \frac{\texttt{1}}{\texttt{r}^{3}}
    \end{equation*}
    \item Charge; pruning
    \begin{equation*}
        \texttt{msg2}=\frac{\texttt{1}}{(\texttt{r}+ \texttt{0.036421545})\texttt{r}^2}\cdot \texttt{q}_1\texttt{q}_2 \cdot(\Delta \texttt{x}-\texttt{0.0331669})+\texttt{0.08641323}
    \end{equation*}
    (shows the correct functional form except there is a small constant added to one of the $1/r$ terms and a $\Delta y$ term is not present).
\end{itemize}


\begin{table}[h!]
\centering
\caption{Pareto front from PySR results on message 2 of pruning model on $r^{-2}$ dataset. The highlighted equation shows the 'best equation' chosen by PySR according to the metric given in Equation \ref{eqn:pysr_score}. The constants have been truncated to two decimal points.}
\begin{tabular}{clc}
\toprule
\textbf{Complexity} & \textbf{Equation} & \textbf{Loss} \\
\midrule
1 & $\phi_2^e = 0.08$ & 0.0888 \\
5 & $\phi_2^e = (\Delta x \cdot 0.00) + 0.08$ & 0.0885 \\
7 & $\phi_2^e = ((\Delta x + \Delta y) \cdot 0.00) + 0.08$ & 0.0882 \\
8 & $\phi_2^e = ((\Delta x \cdot \mathrm{inv}(r)) \cdot 0.00) + 0.08$ & 0.0880 \\
9 & $\phi_2^e = (m_2 \cdot (\Delta y + \Delta x) \cdot 0.00) + 0.08$ & 0.0877 \\
10 & $\phi_2^e = (\mathrm{inv}((r + \Delta x) \cdot r) \cdot -0.00) + 0.08$ & 0.0843 \\
12 & $\phi_2^e = (\mathrm{inv}(r^3) \cdot (\Delta x \cdot 0.02)) + 0.08$ & 0.0687 \\
14 & $\phi_2^e = ((m_2 \cdot 0.01) \cdot (\Delta x \cdot \mathrm{inv}(r^3))) + 0.08$ & 0.0513 \\
16 & $\phi_2^e = ((\Delta y + \Delta x) \cdot ((m_2 \cdot \mathrm{inv}(r^3)) \cdot -0.01)) + 0.08$ & 0.0260 \\
\rowcolor{gray!20}
18 & $\phi_2^e = ((\mathrm{inv}(r^3) \cdot (\Delta x + \Delta y \cdot 0.59)) \cdot m_2) + 0.08$ & 0.0128 \\
20 & $\phi_2^e = (\mathrm{inv}(r^3) \cdot (((m_2 + 0.06) \cdot 0.01) \cdot (\Delta x + \Delta y \cdot 0.59))) + 0.08$ & 0.0125 \\
\bottomrule
\end{tabular}
\label{tab:pareto_front}
\end{table}

\paragraph{Limitations of the framework}
In the reconstructions for the $r^{-1}$ and $r^{-2}$ force laws, the mass of the receiving node, $m_1$, is absent. As described in \cref{eqn:edge_model}, the edge model is intended to learn the interaction forces between particles, with the node model aggregating these messages and outputting the resulting accelerations, as shown in \cref{eqn:edge_model_learns_forces}. However, this setup is mathematically equivalent to having the edge model learn accelerations directly, with the node model simply outputting the aggregated accelerations, as shown in  \cref{eqn:edge_model_learns_acc}.

\textit{Edge model learns forces:}
\begin{align}
    & \phi^e(\textbf{v}_{r_k}, \textbf{v}_{s_k})\approx \text{Force on }r_k \text{ by } s_k \notag\\
    & \phi^v\left(\textbf{v}_{r_k}, \sum_{j\neq r_k} \phi^e (\textbf{v}_{r_k}, \textbf{v}_{j})\right)\approx\phi^v(\textbf{v}_{r_k}, \text{Resultant force on }r_k)\approx \text{Acceleration of }r_k \label{eqn:edge_model_learns_forces}
\end{align}
\textit{Edge model learns accelerations:}
\begin{align}
    & \phi^e(\textbf{v}_{r_k}, \textbf{v}_{s_k})\approx \text{Acceleration on }r_k \text{ caused by interaction with } s_k \notag\\
    & \phi^v\left(\textbf{v}_{r_k}, \sum_{j\neq r_k} \phi^e (\textbf{v}_{r_k}, \textbf{v}_{j})\right)\approx\phi^v(\textbf{v}_{r_k}, \text{Resultant acceleration of }r_k)\approx \text{Acceleration of }r_k \label{eqn:edge_model_learns_acc}
\end{align}

\paragraph{SR Wall Clock Time} To perform SR on the edge model it took approximately 10 minutes on an Apple M4 Max SoC (14-core CPU: 10 performance + 4 efficiency cores, 36\,GB unified memory).

\section{Experimental details and extended results for \cref{subsubsec:chaotic_lorenz}}\label{appndx:chaotic_lorenz}
The code repository for this experiment is located at \url{https://github.com/elizabethsztan/chaotic_lorenz}.
\paragraph{Data generation}
The Lorenz system,
\begin{equation}
    \dot{z}_x = \sigma(z_y - z_x),\qquad
  \dot{z}_y = z_x(\rho - z_z) - z_y,\qquad
  \dot{z}_z = z_x z_y - \beta z_z,
\end{equation}

is integrated with parameters $\sigma = 10$, $\rho = 28$, $\beta = 2.7$ using forward Euler with step size $\Delta t = 0.01$, starting from $z_0 = (0, 1, 1.05)^\top$. A single continuous trajectory of 160\,000 steps is generated and split chronologically into two non-overlapping windows of 80\,000 steps for train and test, respectively. This samples from the natural measure of the Lorenz attractor.

The 3-dimensional Lorenz state is projected into a 10-dimensional observation space via a fixed random linear map $L \in \mathbb{R}^{3 \times 10}$ (entries drawn i.i.d.\ from $\mathcal{N}(0,1)$, seed 290402):
\begin{equation}\label{eqn:chaotic_lift}
    x = zL \in \mathbb{R}^{10}, \qquad \dot{x} = \dot{z}L \in \mathbb{R}^{10}.
\end{equation}
Both the observable states $x$ and their time-derivatives $\dot{x}$ are stored as the training and target data for our experimental pipeline. The ground-truth latent quantities $z, \dot{z}$ are also retained for evaluation but are not used during training.

\paragraph{Model architecture}
The model has three components.

\begin{enumerate}
  \item \textbf{Encoder} $E$: a single learnable affine map $\mathbb{R}^{10} \to \mathbb{R}^3$,
        $h = xW^\top + b$, with weight matrix $W \in \mathbb{R}^{3 \times 10}$ and bias $b \in \mathbb{R}^3$.
  \item \textbf{Dynamics MLP} $f$: a three-layer network with architecture $[3, 64, 64, 3]$ and ReLU activations between hidden layers (no activation on the output). It maps latent states to latent derivatives, $\dot{h} = f(h) \in \mathbb{R}^3$. This block is the target of symbolic distillation.
  \item \textbf{Decoder} $D$: the Moore--Penrose pseudoinverse of the encoder, $D(z) = (z - b)\,(W^\top)^+$. It is recomputed from the encoder weights at each forward pass and has no independent parameters.
\end{enumerate}

The full forward pass is
\begin{equation}
    \hat{\dot{x}} = D(f(E(x))), \qquad \hat{x} = D(E(x)),
\end{equation}
where $\hat{\dot{x}}$ is used for the dynamics loss and $\hat{x}$ is available for reconstruction diagnostics. Because the decoder is the pseudoinverse of the encoder, the bottleneck forces $E$ to find a 3-dimensional coordinate system in which $f$ can accurately predict derivatives in the 10-dimensional observable space.

\paragraph{Training}
The model is trained to minimize
\begin{equation}
      \mathcal{L} = \lambda_{\text{dyn}}\,\mathbb{E}\bigl[\|\dot{x} - \hat{\dot{x}}\|^2\bigr]
              + \lambda_{\text{sparse}}\,\bigl(\mathbb{E}[\|h\|^2] + \mathbb{E}[\|\dot{h}\|^2]\bigr),
\end{equation}
with $\lambda_{\text{dyn}} = 1.0$ and $\lambda_{\text{sparse}} = 10^{-4}$. The sparsity term penalizes the magnitude of latent activations and latent derivatives, encouraging compact representations. Optimization uses Adam with learning rate $10^{-3}$, batch size 8\,000, for 3\,000 epochs. Validation loss (dynamics term only) is monitored every 100 epochs with early stopping: the best checkpoint is restored if loss ceases to improve.

\paragraph{Symbolic distillation}
After training, SymTorch is applied to distill the dynamics MLP $f$ into closed-form symbolic expressions. PySR is configured with binary operators $\{+,\, -,\, \times\}$, no unary operators, maximum expression complexity 35, 400 evolutionary iterations, and optimization probability 0.3. 

\paragraph{Attractor dynamics}
\cref{fig:chaotic_lorenz} demonstrates that the equations discovered by SymTorch recover Lorenz-like attractor dynamics up to a linear coordinate transformation. This transformation corresponds to the learned decoder composed with the inverse of the lift in \cref{eqn:chaotic_lift}. Crucially, since the true lift matrix is unknown at inference time, the method implicitly identifies a lower-dimensional coordinate system in which the governing dynamics take Lorenz form — without requiring access to the ground-truth state coordinates.
\paragraph{Discovered equations}
The discovered equations are 
\begin{align}
    \hat{a}_0 =& (-0.14x_2 + 11.59)(x_2 - x_0) + (0.82x_1 + 4.16)(0.50x_2 + 0.61x_0 + x_1 + 7.09) - 34.48,\\
    \hat{a}_1 =& x_0(-0.34x_1 - 4.46) - (x_1 + 0.53)(0.01x_2^2 - 16.31) + 8.72x_2,\\
    \hat{a}_2 =& (x_0 + 48.96 + 0.04x_1(x_1 + x_2) - x_2)(18.03 - 0.18x_0) - (11.46 - 0.86x_2)x_1 - 884.58.
\end{align}

\paragraph{Finetuning}
A fine-tuning phase trains the encoder/decoder for 200 epochs at learning rate $10^{-4}$ with the symbolic block held fixed, polishing the linear coordinate transform without altering the discovered equations. Fine-tuning is possible because SymTorch's \texttt{switch\_to\_symbolic} replaces the MLP with a differentiable module, so gradients propagate through the symbolic expressions to the encoder parameters.
\paragraph{MSE results on training and test set}
\cref{tab:chaotic_lorenz_train_results} show the MSE performance of the NN, symbolic and finetuned model on the training and test set. Recall that both of these sets come from the same trajectory.
\begin{table}[h]
\centering
\caption{MSE on train and test splits for each model variant on Lorenz trajectory data ($\dot{x}$). Variance is computed over all 10 output dimensions of $\dot{x}$.
}
\begin{tabular}{lcccc}
\toprule
\textbf{Model} & \textbf{Train MSE} & \textbf{Test MSE} &\textit{ Train variance} & \textit{Test variance} \\
\midrule
NN          & 8.22  & 8.18  & \multirow{3}{*}{8646.09} & \multirow{3}{*}{8631.27} \\
Symbolic    & 15.52 & 15.65 &                          &                          \\
Finetune    & 12.51 & 12.66 &                          &                          \\
\bottomrule
\end{tabular}
\label{tab:chaotic_lorenz_train_results}
\end{table}

\paragraph{Comparison with the SINDy autoencoder}
Our setup recreates a benchmark of \citet{champion_2019}: a SINDy autoencoder is used to recover Lorenz dynamics from high-dimensional observations. The key differences are summarized in \cref{tab:sindy_comparison_table}.

\paragraph{SR Wall Clock Time} To perform SR on the NN it took approximately 5-10 minutes on an Apple M4 Max SoC (14-core CPU: 10 performance + 4 efficiency cores, 36\,GB unified memory).

\begin{table}[h]
\centering
\caption{Comparison of the SymTorch method and the SINDy autoencoder method of performing symbolic distillation on high-dimensional data transformed to a lower dimensional latent space.}
\label{tab:sindy_comparison_table}
\begin{tabular}{lll}
\toprule
& \textbf{SINDy autoencoder} & \textbf{Ours (SymTorch)} \\
\midrule
SR method            &\shortstack[l]{Sparse regression \\ over fixed library}  & \shortstack[l]{PySR (genetic algorithm,\\library-free)} \\
\\
Function library     & Must be specified in advance         & Not required \\
\\
Encoder architecture & Unconstrained MLP                    & Single linear layer \\
\\
Latent dynamics      & \shortstack[l]{Linear combination \\ of library terms}  & Arbitrary analytic expressions \\
\bottomrule
\end{tabular}
\end{table}

\section{Experimental details and extended results for \cref{subsubsec:pinns}}\label{appdx:pinns}

\paragraph{PDE systems}
We study three physical systems, each with a known closed-form solution:
\begin{itemize}
    \item{\textbf{1-D Heat Equation PDE}: The temperature field $u(x,t)$ satisfies
\begin{equation}
    \frac{\partial u}{\partial t} = \alpha \frac{\partial^2 u}{\partial x^2}, \quad (x,t) \in [0,1]^2,
\end{equation}
with initial condition $u(x,0) = \sin(\pi x)$ and Dirichlet boundary conditions $u(0,t) = u(1,t) = 0$. The exact solution is
\begin{equation}
    u(x,t) = e^{-\pi^2 \alpha t}\sin(\pi x).
\end{equation}
We set $\alpha = 0.2$ as the ground-truth diffusivity, which is treated as a learnable scalar parameter of the PINN.}
    \item {\textbf{1-D Traveling Wave Equation}: The displacement field $u(x,t)$ satisfies
    \begin{equation}
        \frac{\partial^2 u}{\partial t^2} = c^2 \frac{\partial^2 u}{\partial x^2}, \quad (x,t) \in [0,1]^2,
    \end{equation}
with initial conditions $u(x,0) = \sin(\pi x)$ and $\frac{\partial u}{\partial t}(x,0) = 0$, and Dirichlet boundary conditions $u(0,t) = u(1,t) = 0$. The exact solution is
\begin{equation}
    u(x,t) = \sin(\pi x)\cos(c\pi t).
\end{equation}
We set $c = 0.67$ as the ground-truth wave speed, treated as a learnable scalar parameter of the PINN.}
    \item{\textbf{Damped Harmonic Oscillator}: The displacement $x(t)$ satisfies 
    \begin{equation}
        \frac{d^2 x}{dt^2} + \mu \frac{dx}{dt} + k\, x = 0,
    \end{equation}

with initial conditions $x(0) = 1$ and $x'(0) = 0$, on the domain $t \in [0,1]$. We parameterize with damping coefficient $\delta = \mu/2$ and natural frequency $\omega_0 = \sqrt{k}$. For the under-damped regime ($\delta < \omega_0$), the exact solution is
\begin{equation}
    x(t) = e^{-\delta t}\left[2A\cos(\phi + \omega t)\right],
\end{equation}
where $\omega = \sqrt{\omega_0^2 - \delta^2}$, $\phi = \arctan(-\delta/\omega)$, and $A = 1/(2\cos\phi)$. We use $\delta = 2$ and $\omega_0 = 20$ ($\mu = 4$, $k = 400$), giving $\omega = \sqrt{396} \approx 19.90$. This setup follows \citet{moseley_pinns}.}
\end{itemize}

\paragraph{Data}
For the heat and wave equations, $N = 10$ training points $(x_i, t_i)$ are drawn uniformly at random from $[0,1]^2$, with labels $u_i$ computed from the exact solution. For the damped harmonic oscillator, $N = 10$ training points are taken at uniform spacing from the first portion of the domain, $t \in [0, 0.361]$, covering slightly more than one oscillation period. This tests generalization: the PINN must extrapolate the oscillatory behavior beyond the training window, while the standard NN cannot.

\paragraph{Model architecture}
Both the standard NN and PINN share the same architecture: a fully connected network with 3 hidden layers of 32 units each and Tanh activations. The output layer has a single unit with no activation. For the heat and wave PINNs, the PDE constants ($\alpha$ and $c$, respectively) are treated as additional learnable scalar parameters initialized at 0.0 and 0.5 respectively. No learnable constants are used for the oscillator PINN, as $\mu$ and $k$ are treated as known and hard-coded into the physics residual.

\paragraph{PINN loss formulation}
The PINN loss augments the data-fitting mean squared error with three physics-based penalty terms:

$$\mathcal{L} = \mathcal{L}_{\mathrm{data}} + \lambda_{\mathrm{phys}}\,\mathcal{L}_{\mathrm{phys}} + \lambda_{\mathrm{ic}}\,\mathcal{L}_{\mathrm{ic}} + \lambda_{\mathrm{bc}}\,\mathcal{L}_{\mathrm{bc}},$$

where

$$\mathcal{L}_{\mathrm{phys}} = \frac{1}{N}\sum_{i=1}^{N}\left(\text{differential-equation residual at }(x_i, t_i)\right)^2,$$

and analogously for the initial condition ($\mathcal{L}_{\mathrm{ic}}$) and boundary condition ($\mathcal{L}_{\mathrm{bc}}$) residuals, evaluated at the $N$ training-set time/space coordinates. For the heat and wave equations, $\lambda_{\mathrm{phys}} = 1$, $\lambda_{\mathrm{ic}} = 5$, $\lambda_{\mathrm{bc}} = 5$. For the oscillator, the initial conditions are satisfied implicitly through the training data, so we set $\lambda_{\mathrm{ic}} = \lambda_{\mathrm{bc}} = 0$ and $\lambda_{\mathrm{phys}} = 10^{-4}$; the physics residual is evaluated on 30 fixed collocation points spanning $[0,1]$.
Derivatives for the physics residuals are computed via PyTorch's automatic differentiation.

\paragraph{Training}

All models are trained with Adam. For the heat equation, we use learning rate $10^{-3}$ for 3000 epochs. For the wave equation, we use learning rate $10^{-3}$ for 5000 epochs. For the oscillator, the standard NN uses learning rate $10^{-3}$ for 20000 epochs, while the PINN uses learning rate $10^{-4}$ for 20000 epochs. All experiments use random seed 290402.

\paragraph{Symbolic distillation with SymTorch}

After training, we apply SymTorch to distill each PINN into a closed-form symbolic expression. For each system, we generate 5000 sample points drawn uniformly from the input domain and pass them through the trained PINN to obtain input--output pairs. These are then used as the dataset for PySR. Variable names are set to $[\texttt{x}, \texttt{t}]$ for the PDEs and $[\texttt{t}]$ for the oscillator.

The SR parameters are common across all three systems: unary operators $\{\texttt{inv}(\cdot) = 1/x,\, \texttt{sin},\, \texttt{exp}\}$ (with $\texttt{cos}$ added for the wave equation and oscillator), complexity of operators $\{\texttt{sin}: 3,\, \texttt{cos}: 3,\, \texttt{exp}: 3\}$, parsimony $= 0.01$, and 1000 evolutionary iterations. Nested constraints prevent compositions such as $\sin(\sin(\cdot))$ and $\exp(\exp(\cdot))$. 

\paragraph{Results}

The symbolic expressions recovered at complexity 13 are shown in \cref{tab:pinn_results}, alongside the true solutions and the values of any recovered constants.

\begin{table}[h]
\centering
\caption{Symbolic expressions recovered by SymTorch at complexity 13 for each physical system, alongside the true closed-form solutions and recovered physical constants.}
\label{tab:pinn_results}
\begin{tabular}{l >{\normalsize}l >{\tiny}l >{\tiny}l >{\tiny}l}
\toprule
\textbf{System }
& \normalsize \textbf{True solution}
& \normalsize \textbf{Distilled expression}
& \normalsize \textbf{True constant}
& \normalsize \textbf{SR estimate} \\
\midrule
Heat equation
& $e^{-\pi^2\alpha t}\sin(\pi x)$
& $\sin(3.144\, x)\cdot e^{-1.981\, t}$
& $\alpha = 0.200$
& $\substack{\hat{\alpha} = 1.981/\pi^2\\ \approx 0.201}$ \\
\\
Wave equation
& $\sin(\pi x)\cos(c\pi t)$
& $\sin(3.147\, x)\cdot\cos(2.105\, t)$
& $c = 0.670$
& $\substack{\hat{c} = 2.105/\pi\\ \approx 0.670}$ \\
\\
Oscillator
& $e^{-\delta t}\cdot 2A\cos(\phi + \omega t)$
& $e^{-2.000\, t}\cdot\cos(-19.625\, t)$
& $\delta=2,\;\omega\approx19.90$
& $\substack{\hat{\delta} = 2.000\\ \hat{\omega} \approx 19.625}$ \\
\bottomrule
\end{tabular}
\end{table}

For all three systems, the distilled expression recovers the correct functional form of the exact PDE solution. In the heat and wave cases, the coefficient of $\pi$ is recovered to within 0.1\%, and the recovered PDE constant matches the ground truth to three significant figures. For the oscillator, the damping coefficient $\delta = 2$ is recovered exactly, and the oscillation frequency $\hat{\omega} \approx 19.625$ is close to the true value $\sqrt{396} \approx 19.900$. The small phase offset $\phi \approx -0.100$ is not recovered, as it is absorbed into the argument of $\cos$; the SR did not manage to recover this precisely.\\
The heat and wave PINNs also recover accurate estimates of the unknown PDE/ODE constants directly from the trained model parameters ($\hat{\alpha} = 0.196$ vs.\ true $0.200$; $\hat{c} = 0.666$ vs.\ true $0.670$), demonstrating that PINNs can jointly learn the solution and infer unknown physical parameters from sparse data.

\paragraph{SR Wall Clock Time} To perform SR on each PINN it took approximately 5 minutes on an Apple M4 Max SoC (14-core CPU: 10 performance + 4 efficiency cores, 36\,GB unified memory).

\section{SLIME and LIME benchmarking}\label{appndx:slime_lime_benchamrking}
The source code for this benchmarking is available at \url{https://github.com/elizabethsztan/SLIMEvLIME_benchmarks}.
\subsection{Experimental setup}
\paragraph{Datasets}
We use six classification and two regression datasets sourced from Scikit-learn and OpenML, as described in \cref{tab:slime_lime_dataset}. The number of nearest neighbors used for local variance estimation, $J$, is set to $\min(100, \max(10, \text{floor}(0.05 × n_{\text{train}})))$. This heuristic selects a neighborhood that scales with the dataset: large enough to give a stable variance estimate, but small enough to remain genuinely local.

\begin{table}[ht]
\centering
\caption{Summary of datasets used in the SLIME and LIME benchmarking.}
\label{tab:slime_lime_dataset}
\begin{tabular}{llrrrrr}
\toprule
Dataset & Task & $n_{\mathrm{train}}$ & $n_{\mathrm{test}}$ & $n_{\mathrm{features}}$ & $J$ & $n_{\mathrm{syn}}$ \\
\midrule
Iris & Classification & 120 & 30 & 4 & 10 & 500 \\
Wine & Classification & 142 & 36 & 13 & 10 & 500 \\
Breast Cancer & Classification & 455 & 114 & 30 & 22 & 500 \\
Pima Diabetes & Classification & 614 & 154 & 8 & 30 & 500 \\
German Credit & Classification & 800 & 200 & 61 & 40 & 500 \\
Adult & Classification & 39,073 & 9,769 & 108 & 100 & 500 \\
Concrete & Regression & 824 & 206 & 8 & 41 & 500 \\
California Housing & Regression & 16,512 & 4,128 & 8 & 100 & 500 \\
\bottomrule
\end{tabular}
\end{table}

\paragraph{Black-box model} For each dataset we train a fully-connected MLP with two hidden layers of width 64 (classification) or 128 (regression), ReLU activations, and Adam optimization for between 100 to 200 epochs. Test-set performance ranges from 70\% accuracy (Pima Diabetes) to 100\% (Iris) for classification, and $R^2$ of 0.79 (California Housing) to 0.89 (Concrete) for regression.

\paragraph{Local neighborhood} For each of five randomly sampled local points $x$, we draw $n_{\text{syn}} = 500$ synthetic samples by perturbing $x$ with Gaussian noise whose per-feature variance is estimated from the $J$ nearest training-set neighbors. Kernel weights for neighborhood-averaged metrics use the Gaussian kernel $w(z) = \exp(-\lVert z - x\rVert^2 / \sigma^2)$, where $\sigma^2$ is the same neighborhood variance. Each surrogate is then evaluated on a fresh set of 500 held-out points drawn from the same local distribution.

\paragraph{LIME and SLIME fitting} We use the Lime Python library \cite{lime} with default settings. To fit SLIME, we use SymTorch's \texttt{SymbolicModel} in SLIME mode using default hyperparameters, with the number of PySR iterations increased to 600.

\paragraph{Metrics}
The metrics used to compare SLIME and LIME surrogate models are
\begin{itemize}
    \item Classification: kernel-weighted cross-entropy (WCE), kernel-weighted probability MSE (WP-MSE), kernel-weighted decision-boundary agreement (Agree.), and KL divergence between surrogate and model predictions at the exact local point $x$.
    \item Regression: kernel-weighted $R^2$ (W$\text{R}^2$), kernel-weighted MAE (WMAE), and kernel-weighted MSE (WMSE). WMAE, WMSE, predictions at the local point $x$, and regression equations are reported in standardized target units rather than original target units.
\end{itemize}

\begin{table}[h]
\centering
\tiny
\caption{SLIME vs LIME results: Adult.}
\label{tab:adult}
\begin{tabular}{cccccccccc}
\toprule
      & \multicolumn{4}{c}{LIME} & \multicolumn{5}{c}{SLIME} \\
\cmidrule(lr){2-5} \cmidrule(lr){6-10}
Point index & KL div & WCE & WP-MSE & Agree. & KL div & WCE & WP-MSE & Agree. & Example equation for prob. of class 0 \\
\midrule
4116 & 0.73 & 0.61 & 0.01 & 1.00 & 0.23 & 0.61 & 0.00 & 1.00 &
\shortstack[l]{
$\operatorname{inv}(x_{24} + (x_{33}+x_{25})(x_{38}$\\
$+ 32.90(x_{48}+x_{47}-1.03(x_{24}+$\\$x_{49})(x_{11}+x_{61}))) + 1.06)$} \\
\\
11317 & 0.95 & 1.17 & 0.89 & 0.05 & 0.00 & 1.69 & 0.08 & 0.95 &
\shortstack[l]{
$\operatorname{inv}(3.08((20.24(x_{58}+x_{24}+x_{41}$\\
$+x_{51}+x_{10}+x_{61}) + x_9)$\\
$(x_{53}+x_{23}) + 3.10))$} \\
\\
11038 & 0.31 & 0.48 & 0.29 & 1.00 & 0.00 & 0.10 & 0.02 & 1.00 &
\shortstack[l]{
$-0.79(((x_{24}+0.21+x_{25}+x_{47})$\\
$(\exp(x_{40})x_{33}+x_{67})$\\
$- 1.25 + x_{30}x_{41})$} \\
\\
184 & -0.00 & 0.00 & 0.00 & 1.00 & 0.68 & 0.27 & 0.11 & 1.00 &
\shortstack[l]{
$\operatorname{inv}(x_2(x_{25}+x_6+x_{97}$\\
$+0.03x_{33}+x_{27})$\\
$+ x_{11} + 0.99)$} \\
\\
30797 & 0.38 & 1.17 & 0.70 & 0.32 & 0.04 & 15.94 & 0.98 & 0.32 &
\shortstack[l]{
$\operatorname{inv}(((x_{24}+x_{38}+0.34)x_{33}x_{63} $\\$+ 1.71)(x_{21}x_{31}+0.41)$\\
$+ x_{10}+x_{29}+x_{49}x_{30}-0.46)$} \\
\bottomrule
\end{tabular}
\end{table}

\begin{table}[h]
\centering
\tiny
\caption{SLIME vs LIME results: Breast cancer. $\epsilon$ symbolizes a small constant.}
\label{tab:breast_cancer}
\begin{tabular}{cccccccccc}
\toprule
      & \multicolumn{4}{c}{LIME} & \multicolumn{5}{c}{SLIME} \\
\cmidrule(lr){2-5} \cmidrule(lr){6-10}
Point index & KL div & WCE & WP-MSE & Agree. & KL div & WCE & WP-MSE & Agree. & Example equation for prob. of class 0 \\
\midrule
47 & 0.66 & 0.77 & 0.56 & 0.09 & 0.00 & 0.04 & 0.00 & 1.00 &
\shortstack[l]{$x_{10}\operatorname{inv}(x_9)(x_{12}((x_{28}x_1 $\\$- 3.04)(\epsilon x_{26}+ \epsilon (x_1 -$\\$ 14.92))) + \epsilon)$} \\
\\
130 & 1.03 & 1.22 & 0.89 & 0.00 & 0.01 & 0.31 & 0.05 & 1.00 &
\shortstack[l]{$(x_{27} - 0.04\sin(x_{21}))((-\epsilon x_{23} - $\\$0.11)(x_3 - 0.09x_{21}^2) + $\\$17.25) + 0.19$} \\
\\
128 & 0.00 & 0.01 & 0.00 & 1.00 & 0.00 & 0.01 & 0.00 & 1.00 &
\shortstack[l]{$\operatorname{inv}(x_1)^2x_3^2(-\epsilon) + 1.00$\\
$+ x_{19}\operatorname{inv}(\operatorname{inv}(x_7)(\sin(x_{20}) $\\$- 0.15))$} \\
\\
2 & 0.04 & 0.07 & 0.00 & 1.00 & 0.00 & 0.04 & 0.00 & 1.00 &
\shortstack[l]{$x_6(x_{16} - 0.05x_{11})$\\
$+ x_{21}(1.75(x_{11}- $\\$ 1.37)x_{11}x_{14}^2 + \epsilon )$} \\
\\
357 & 3.42 & 4.18 & 0.06 & 1.00 & 0.00 & 3.75 & 0.05 & 1.00 &
\shortstack[l]{$x_1(-\epsilon x_{13} + \epsilon x_1)$\\
$+ 6.91x_{28}(\epsilon x_{23} + x_6 - 0.03x_2)$} \\
\bottomrule
\end{tabular}
\end{table}

\begin{table}
\centering
\tiny
\caption{SLIME vs LIME results: California Housing. $\epsilon$ symbolizes a small constant.}
\label{tab:california}
\begin{tabular}{cccccccc}
\toprule
      & \multicolumn{3}{c}{LIME} & \multicolumn{4}{c}{SLIME} \\
\cmidrule(lr){2-4} \cmidrule(lr){5-8}
Point index & WR$^2$ & WMAE & WMSE & WR$^2$ & WMAE & WMSE & Example equation \\
\midrule
1739 & -39.59 & 4.90 & 24.30 & -0.17 & 0.67 & 0.70 &
\shortstack[l]{$-0.36x_2 + 1.43x_0\operatorname{inv}(x_5)$} \\
\\

4782 & -74.08 & 4.21 & 18.47 & 0.21 & 0.35 & 0.19 &
\shortstack[l]{$16.22x_6\operatorname{inv}(x_7 + 79.84)$\\
$+ x_0\operatorname{inv}(x_5) + 13.07$} \\
\\

4664 & -6.64 & 2.95 & 9.02 & 0.81 & 0.36 & 0.22 &
\shortstack[l]{$3.00(\operatorname{inv}(x_5)+\operatorname{inv}(x_2))$\\
$+ ((-\epsilon x_7 - 0.35)x_6 + 3.35)x_0 - 3.37$\\
$+ \operatorname{inv}(x_6 - 2.27x_0)$} \\
\\

77 & -30.12 & 2.96 & 9.40 & -0.07 & 0.41 & 0.32 &
\shortstack[l]{$-0.35x_2 + (x_0 + \sin(x_2))$\\
$(\operatorname{inv}(x_5) + 0.15)$} \\
\\

13014 & -31.57 & 3.28 & 11.10 & -0.18 & 0.48 & 0.40 &
\shortstack[l]{$x_2(0.08x_0x_3 - 0.59)$\\
$+ \operatorname{inv}(x_5(\operatorname{inv}(x_1) + 0.24))$} \\
\bottomrule
\end{tabular}
\end{table}

\begin{table}
\centering
\tiny
\caption{SLIME vs LIME results: Concrete. $\epsilon$ symbolizes a small constant.}
\label{tab:concrete}
\begin{tabular}{cccccccc}
\toprule
      & \multicolumn{3}{c}{LIME} & \multicolumn{4}{c}{SLIME} \\
\cmidrule(lr){2-4} \cmidrule(lr){5-8}
Point index & WR$^2$ & WMAE & WMSE & WR$^2$ & WMAE & WMSE & Example equation \\
\midrule
86 & -22.94 & 1.28 & 1.68 & 0.21 & 0.19 & 0.06 &
\shortstack[l]{$(-\epsilon x_7 + 0.04)x_7 + 0.01x_0$\\
$+ x_3(-\epsilon x_4^2 - 0.02)$} \\
\\

237 & -13.86 & 1.85 & 3.49 & 0.62 & 0.26 & 0.09 &
\shortstack[l]{$-0.04x_0\operatorname{inv}(x_3\operatorname{inv}(-0.04x_2x_7 - 0.44x_1) $\\$- 1.34) - 2.53$} \\
\\

232 & -54.81 & 1.38 & 2.00 & -0.74 & 0.20 & 0.06 &
\shortstack[l]{$(\operatorname{inv}(x_3)x_6 - 2.14)(-\epsilon x_6 + 4.04)$\\
$((\operatorname{inv}(x_3)x_6 - 3.58)(-0.01x_6 + 6.99) + 1.56)$} \\
\\

3 & 0.69 & 0.18 & 0.05 & 0.81 & 0.13 & 0.03 &
\shortstack[l]{$(0.07x_1 + 5.10)(\operatorname{inv}(x_3)x_7 - 0.47$\\
$+ \epsilon (x_1 + x_0) + \epsilon x_2)$} \\
\\

648 & -9.38 & 1.02 & 1.15 & 0.22 & 0.23 & 0.09 &
\shortstack[l]{$\operatorname{inv}(x_7 + 55.98)(x_2\sin(x_3) - 166.70)$\\
$+ (x_0 + 17.89x_4)(\operatorname{inv}(x_3) - \epsilon )$} \\
\bottomrule
\end{tabular}
\end{table}

\begin{table}[h]
\centering
\tiny
\caption{SLIME vs LIME results: German Credit.}
\label{tab:german_credit}
\begin{tabular}{cccccccccc}
\toprule
      & \multicolumn{4}{c}{LIME} & \multicolumn{5}{c}{SLIME} \\
\cmidrule(lr){2-5} \cmidrule(lr){6-10}
Point index & KL div & WCE & WP-MSE & Agree. & KL div & WCE & WP-MSE & Agree. & Example equation for prob. of class 0 \\
\midrule
83 & 1.48 & 1.65 & 0.01 & 1.00 & 0.01 & 0.24 & 0.00 & 1.00 &
\shortstack[l]{$-0.17x_{60}(\operatorname{inv}(x_2) $\\$+x_{57}(2.59x_8 + (x_{18} $\\$+x_{20})(x_7+x_{54})) +$\\$ x_{12}(x_{52} - 0.60)) + 0.90$} \\
\\
230 & 0.01 & 0.39 & 0.01 & 1.00 & 0.00 & 0.76 & 0.01 & 1.00 &
\shortstack[l]{$0.04(x_{32}+$\\$x_7+x_{31}+x_{51} +$\\$ (x_{58}+x_{25})(x_{34}+$\\$0.21x_3))+ $\\$0.07x_{10} + 0.82$} \\
\\
225 & 0.06 & 0.85 & 0.15 & 0.05 & 0.00 & 0.74 & 0.05 & 0.05 &
\shortstack[l]{$0.05(x_{56}+x_{41}) - $\\$0.10(x_{57}(x_8+x_{52}$\\$+x_{11}+x_{38})+ (x_7+$\\$x_{11})\operatorname{inv}(x_3)) + 0.95$} \\
\\
3 & 3.42 & 5.14 & 0.07 & 1.00 & 0.00 & 0.48 & 0.00 & 1.00 &
\shortstack[l]{$-0.07(x_8 + (x_{53}+$\\$x_{47})(x_{51}+x_{35}+$\\$x_{33}))(x_{15}+$\\$x_{28})+ 0.96 - $\\$0.20x_{15}x_{50}$} \\
\\
629 & 1.63 & 2.52 & 0.02 & 1.00 & 0.00 & 0.31 & 0.00 & 1.00 &
\shortstack[l]{$2.67\operatorname{inv}((x_{45}+$\\$0.60)((0.69(x_8+$\\$0.22x_{54})+$\\
$x_{26})(x_{57}+$\\$x_{20}+x_{33})) + $\\$x_{15}+2.81)$} \\
\bottomrule
\end{tabular}
\end{table}

\begin{table}
\centering
\tiny
\caption{SLIME vs LIME results: Iris. $\epsilon$ symbolizes a small constant.}
\label{tab:iris}
\begin{tabular}{cccccccccc}
\toprule
      & \multicolumn{4}{c}{LIME} & \multicolumn{5}{c}{SLIME} \\
\cmidrule(lr){2-5} \cmidrule(lr){6-10}
Point index & KL div & WCE & WP-MSE & Agree. & KL div & WCE & WP-MSE & Agree. & Example equation for prob. of class 0 \\
\midrule
12 & 0.00 & 0.04 & 0.00 & 1.00 & 0.00 & 0.02 & 0.00 & 1.00 &
\shortstack[l]{$(\exp(-x_0-x_2))(x_3 +$\\
$ 2.60)(\operatorname{inv}(x_2) -$\\$ 0.07)\exp(x_1)$} \\
\\
33 & 1.40 & 1.42 & 1.14 & 0.00 & 0.00 & 0.02 & 0.00 & 1.00 &
\shortstack[l]{$x_2((x_3x_2^2 +$\\$ 0.17)x_2(\epsilon x_0$\\
$- \epsilon x_2^2\operatorname{inv}(x_1 - $\\$ 2.78))) + 1.00$} \\
\\
119 & 0.66 & 0.69 & 0.48 & 0.48 & 0.00 & 0.06 & 0.00 & 1.00 &
\shortstack[l]{$\epsilon \sin(x_0)(-0.49x_2 +$\\$ x_1)(\operatorname{inv}(x_3 + \sin(x_2) - $\\$ 0.98) - 1.20)$} \\
\\
0 & 0.41 & 0.41 & 0.23 & 1.00 & -0.00 & 0.00 & 0.00 & 1.00 &
\shortstack[l]{$-\epsilon \operatorname{inv}(\operatorname{inv}(\operatorname{inv}(0.31x_0 - $\\$ 3.50x_3^2)x_2^2 + 0.18)$\\
$+ x_1 - 3.50) + 1.00$} \\
\\
93 & 0.05 & 0.36 & 0.02 & 1.00 & 0.00 & 0.30 & 0.00 & 1.00 &
\shortstack[l]{$\epsilon (x_1+x_0)(x_1 - 5.92)x_2$\\
$+ \epsilon$} \\
\bottomrule
\end{tabular}
\end{table}

\begin{table}
\centering
\tiny
\caption{SLIME vs LIME results: Pima Diabetes. $\epsilon$ symbolizes a small constant.}
\label{tab:pima_diabetes}
\begin{tabular}{cccccccccc}
\toprule
      & \multicolumn{4}{c}{LIME} & \multicolumn{5}{c}{SLIME} \\
\cmidrule(lr){2-5} \cmidrule(lr){6-10}
Point index & KL div & WCE & WP-MSE & Agree. & KL div & WCE & WP-MSE & Agree. & Example equation for prob. of class 0 \\
\midrule
64 & 0.17 & 0.70 & 0.15 & 0.34 & 0.01 & 0.55 & 0.00 & 0.99 &
\shortstack[l]{$(0.02x_7^2 + 2.99x_0x_6 +$\\$ x_5)\operatorname{inv}(x_2)- 0.39$} \\
\\
176 & 0.51 & 0.77 & 0.43 & 0.04 & 0.00 & 0.30 & 0.00 & 1.00 &
\shortstack[l]{$\epsilon x_1 + (-0.01x_5 + $\\$0.04x_0x_6)$\\
$(x_0+x_2+x_3)\operatorname{inv}(x_4)$} \\
\\
173 & 0.02 & 0.68 & 0.03 & 0.91 & 0.00 & 0.66 & 0.02 & 0.98 &
\shortstack[l]{$((x_6((x_6x_1 - 85.14)x_6) +$\\$ x_0)\epsilon \sin(x_0)$\\
$- 0.01)x_2 + 0.68$} \\
\\
2 & 0.19 & 0.73 & 0.19 & 0.34 & 0.01 & 0.56 & 0.03 & 0.97 &
\shortstack[l]{$\epsilon(\operatorname{inv}(\sin(x_5) - $\\$0.71) + x_4 + x_5+x_1$\\
$+ \operatorname{inv}(x_6 - 0.12)) - 0.45$} \\
\\
483 & 0.90 & 1.20 & 0.76 & 0.00 & 0.00 & 0.38 & 0.01 & 1.00 &
\shortstack[l]{$x_4((-\epsilon (x_2+x_7+x_2) +$\\$ \epsilon)x_1$\\
$- 0.01)$} \\
\bottomrule
\end{tabular}
\end{table}

\begin{table}
\centering
\tiny
\caption{SLIME vs LIME results: Wine. $\epsilon$ symbolizes a small constant.}
\label{tab:wine}
\begin{tabular}{cccccccccc}
\toprule
      & \multicolumn{4}{c}{LIME} & \multicolumn{5}{c}{SLIME} \\
\cmidrule(lr){2-5} \cmidrule(lr){6-10}
Point index & KL div & WCE & WP-MSE & Agree. & KL div & WCE & WP-MSE & Agree. & Example equation for prob. of class 0 \\
\midrule
14 & 0.38 & 0.79 & 0.25 & 1.00 & 0.00 & 0.34 & 0.00 & 1.00 &
\shortstack[l]{$6.15(\operatorname{inv}(x_2x_4x_3)(x_8 + $\\$0.03x_{12})$\\
$- \epsilon + \epsilon x_5)$} \\
\\
39 & 1.04 & 1.52 & 0.06 & 1.00 & 0.00 & 0.25 & 0.00 & 1.00 &
\shortstack[l]{$\epsilon x_8 + \operatorname{inv}((5.23x_7 +$\\$ x_3)x_2)$\\
$+ \epsilon x_{12} - 0.04$} \\
\\
40 & 0.28 & 0.52 & 0.13 & 1.00 & 0.00 & 0.23 & 0.00 & 1.00 &
\shortstack[l]{$((x_3 + \operatorname{inv}(x_1)x_5x_{11})x_{10} + $\\$x_4)
(-1.86\operatorname{inv}(x_{12})) + 1.20$} \\
\\
0 & 0.24 & 0.45 & 0.03 & 1.00 & 0.00 & 0.24 & 0.00 & 1.00 &
\shortstack[l]{$\operatorname{inv}((x_1+\sin(x_3))$\\$(0.02\exp(x_9))$\\
$+ x_7 + 1.99) + 0.02$} \\
\\
111 & 2.65 & 3.54 & 0.07 & 1.00 & 0.00 & 0.57 & 0.00 & 1.00 &
\shortstack[l]{$x_5^4((-0.01x_4 + 0.87)(-0.01x_4 $\\$+ 0.78))
+ 0.02x_8x_{10}$} \\
\bottomrule
\end{tabular}
\end{table}

\clearpage
\section{Experimental details and extended results for \cref{subsec:llm_interp}}\label{appdx:llm_operations}
\subsection{Setup}
The specific prompts used to query the Llama-3.2-1B-Instruct model are shown in \cref{tab:llm_operations_prompts}. When generating text, we configured the LLM to perform greedy decoding (\texttt{do\_sample=False}) with a maximum generation limit of 250 new tokens. Since the LLM responses included explanatory text beyond the numeric answer (eg. step-by-step reasoning), we implemented a regex-based extraction function to parse the final numeric result from the LLM's output. We then created a wrapper function for each mathematical operation (addition, multiplication, temperature conversion, and counting) that: (1) took the numerical inputs as a NumPy array (eg. pairs of 3-digit integers for addition), (2) formatted these inputs into natural language prompts, (3) queried the LLM, (4) extracted the numeric answer, and (5) returned the result as a NumPy array. Finally, we used SymTorch's model-agnostic mode to wrap these functions and perform SR on the LLM's input-output behavior. 

The SR parameters used are the default ones as in \cref{tab:def_sr_params} but we added a constraint on $\sin$ and $\exp$ to only take arguments of up to complexity 1 and ran the SR for 5000 iterations.

\begin{table}[ht]
\caption{Prompts used to query Llama-3.2-1B-Instruct on four mathematical tasks: addition, multiplication, Celsius-to-Fahrenheit conversion, and counting 1s in binary strings. All prompts request answers in \texttt{\textbackslash boxed\{\}} format to enable regex-based extraction of numeric results. For the temperature conversion task, we only inputted temperatures between $-20^\circ\mathrm{C}$ and $200^\circ\mathrm{C}$.}
  \label{tab:llm_operations_prompts}
  \centering
  \begin{small}
    \begin{tabular}{cc}
      \toprule
      Operation &
      Example Prompt \\
      \midrule
      Addition of two 3-digit numbers &
      \shortstack{Return only the numeric answer in the format \$boxed\$.\\What is $193+374=$?} \\
      \midrule
      Multiplication of two 3-digit numbers &
      \shortstack{Return only the numeric answer in the format \$boxed\$.\\What is $484 * 726=$?} \\
      \midrule
      Counting 1s in a 6-digit string of 1s and 0s &
      \shortstack{Return only the numeric answer in the format \$boxed\$.\\How many 1s are there in the string 000101?}\\
      \midrule
      Converting from Celsius to Fahrenheit &
      \shortstack{Return only the numeric answer in the format \$boxed\$.\\What is $30$ degrees Celsius in Fahrenheit?}\\
      \bottomrule
    \end{tabular}
  \end{small}
  \vskip -0.1in
\end{table}

\paragraph{SR Wall-Clock Time} Symbolic distillation required 5-15 minutes on an Apple M4 Max SoC (14-core CPU: 10 performance + 4 efficiency cores, 36\,GB unified memory) for each of these operations.

\clearpage
\section{Experimental details and extended results for \cref{sec:symdistill}}\label{adx:symdistill}
\begin{figure}[h]
    \centering
    \includegraphics[width=0.85\linewidth]{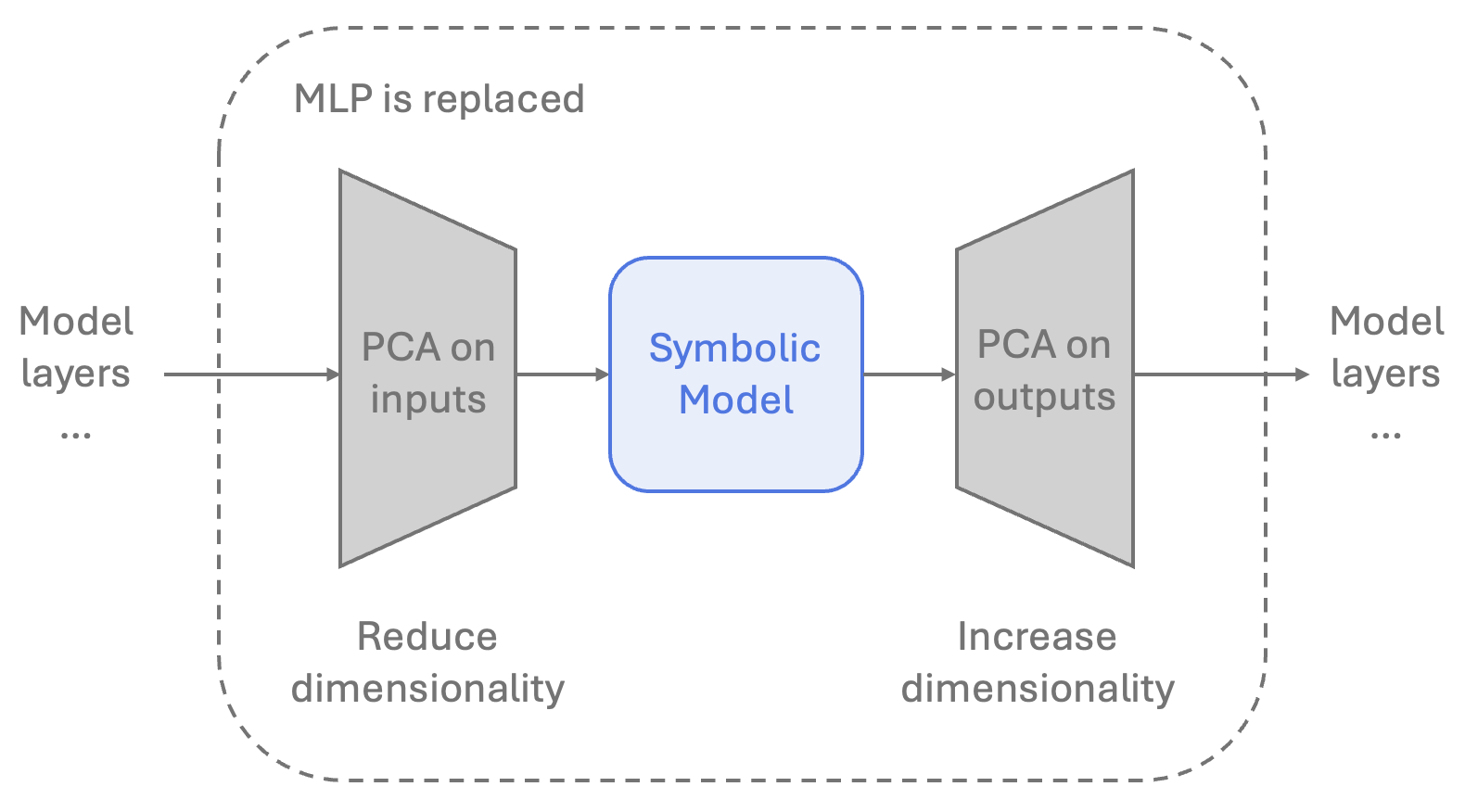}
    \caption{Conceptual overview of the symbolic distillation pipeline applied to an LLM MLP layer. Input activations are projected to a reduced PCA subspace, symbolic regression learns a compact mapping between the compressed input and output representations, and the resulting equations replace the full MLP during inference.}
    \label{fig:llm_pca_conceptual}
\end{figure}

The code repository for this experiment is located at \url{https://github.com/elizabethsztan/LLM_acceleration_SymTorch}.
\paragraph{Conceptual overview}
\cref{fig:llm_pca_conceptual} illustrates the pipeline. For each MLP layer, input activations are compressed to a low-dimensional PCA subspace, symbolic regression learns a compact mapping between the compressed input and output representations, and the resulting equations replace the full MLP in subsequent forward passes.

\paragraph{Setup}
We used the Qwen2.5-1.5B-Instruct model \citep{qwen2.5} for these experiments and the Wikitext-2-v1 dataset \citep{wikitext}. Training and validation sets each consisted of 178k tokens (a randomly selected chunk of 750k characters from each split). The training set was used to fit both the PCA models and the SR surrogates. Perplexity on the held-out test set quantifies model quality:
\begin{equation}
    \text{Perplexity} = \exp\!\left(-\frac{1}{N}\sum_{i=1}^{N}\log p(x_i \mid x_{<i})\right).
\end{equation}
When computing perplexity over the dataset, a chunk size of 1048 tokens was used.

\paragraph{Per-layer sensitivity}
\cref{tab:per_layer_ppl} reports the individual perplexity impact of a PCA-only and a PCA+SR intervention on each of the 28 MLP layers, measured independently on the test set. Layers are ranked from lowest to highest perplexity impact; greedy selection turns on surrogates in this order.

\begin{table}[t]
  \centering
  \caption{Per-layer perplexity delta ($\Delta$PPL) for PCA and PCA+SR interventions. The validation set was used to produce these results.}
  \label{tab:per_layer_ppl}
  \small
  \setlength{\tabcolsep}{6pt}
  \begin{tabular}{
    c
    S[table-format=4.4]
    S[table-format=6.4]
  }
    \toprule
    {Layer} & {PCA} & {PCA+SR} \\
    \midrule
    0  &   89.5918 &   106.6565 \\
    1  &    4.8133 & 1981.6998 \\
    2  &    0.8865 &     0.8842 \\
    3  &    0.7172 &     0.7848 \\
    4  &    0.8810 &     0.8346 \\
    5  &    0.9624 &     0.9588 \\
    6  &    0.7994 &     0.8022 \\
    7  &    1.0050 &     0.9922 \\
    8  &    0.8667 &     0.8275 \\
    9  &    0.5671 &     0.5150 \\
    10 &    0.4231 &     0.4439 \\
    11 &    0.5714 &     0.5458 \\
    12 &    0.5577 &     0.5056 \\
    13 &    0.6662 &     0.4550 \\
    14 &    0.4796 &     0.4558 \\
    15 &    0.4511 &     0.4409 \\
    16 &    0.4434 &     0.4708 \\
    17 &    0.4339 &     0.4414 \\
    18 &    0.5824 &     0.5906 \\
    19 &    0.9160 &     0.9141 \\
    20 &    0.6380 &     0.6391 \\
    21 &    1.0043 &     1.0197 \\
    22 &    1.2190 &     1.2070 \\
    23 &    1.1381 &     1.1692 \\
    24 &    1.0192 &     1.0456 \\
    25 &    1.1992 &     1.2606 \\
    26 &    1.2251 &     1.1790 \\
    27 &    5.8876 &     6.3248 \\
    \bottomrule
  \end{tabular}
\end{table}

\paragraph{PCA sensitivity analysis}
Before performing SR, we determine the minimum number of principal components needed to preserve model behavior. We vary the number of PCA components for both the input and output activations and measure the resulting perplexity change. Input activations are projected to a lower-dimensional subspace via PCA and reconstructed before the MLP; outputs are similarly projected and reconstructed before being passed to the remainder of the model. The analysis was performed by adding PCA to layers 7, 14, and 21, chosen as they were evenly spaced layers. The original SwiGLU MLP projects inputs from a 1536-dimensional space to 8960 dimensions and back, making the chosen 32$\to$8 subspace a substantial compression.

\begin{figure}[h]
    \centering
    \includegraphics[width=1\linewidth]{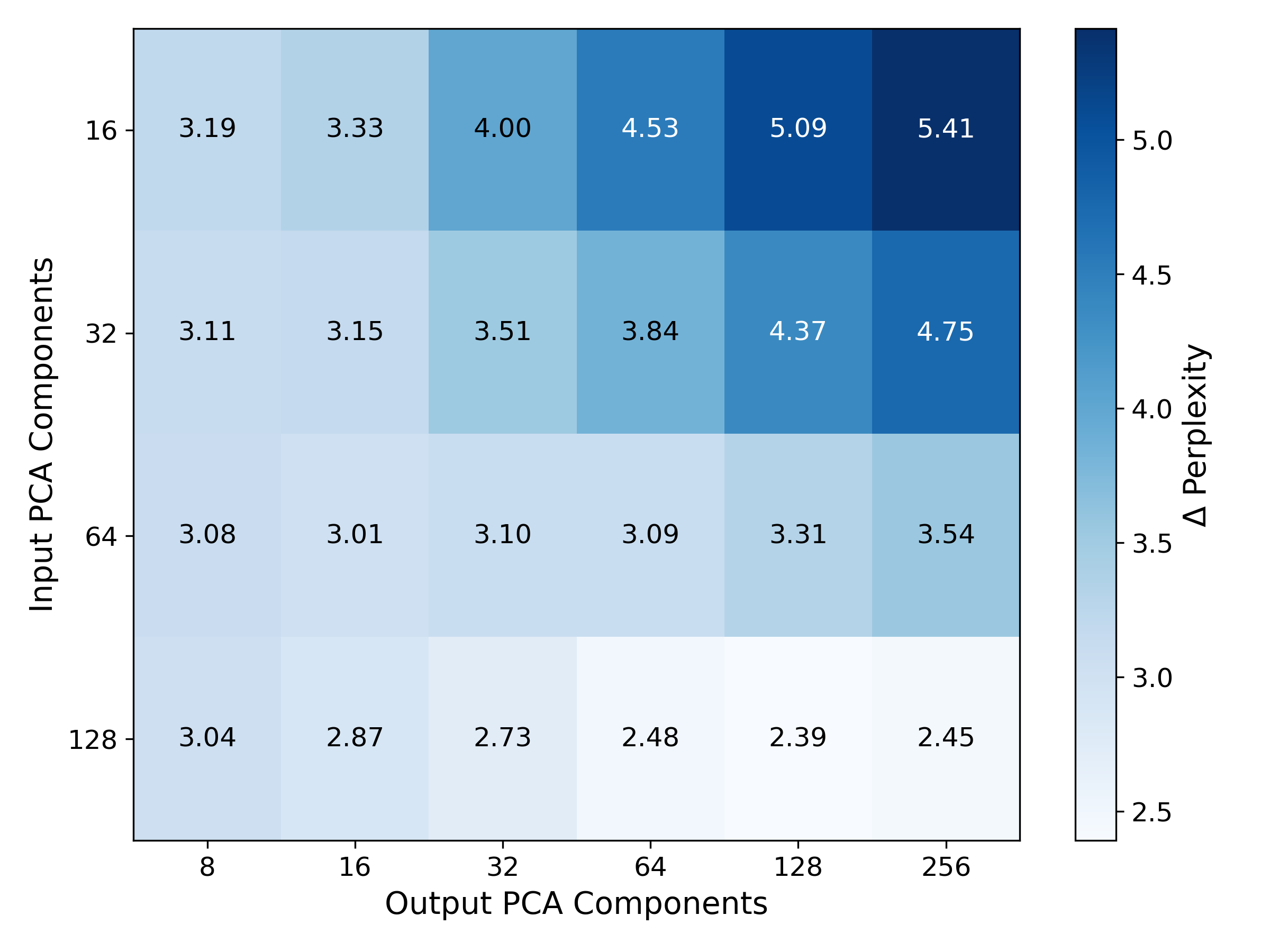}
    \caption{Change in validation set perplexity under PCA compression and reconstruction of MLP activations, relative to the baseline perplexity of 10.62. Input activations are projected to the number of components shown on the $x$-axis; output activations to the number shown on the $y$-axis. The analysis was performed by adding PCA to the arbitrarily selected layers 7, 14, and 21.}
    \label{fig:pca_results}
\end{figure}

\cref{fig:pca_results} shows the sensitivity results. We select 32 principal components for the MLP inputs and 8 for the outputs. Notably, for a fixed number of input PCA components, perplexity initially decreases as the number of output components increases before rising again. A possible explanation is that moderate output compression removes low-variance or noisy directions while preserving the dominant functional subspace, whereas too many retained components reintroduce poorly conditioned directions that degrade downstream representations.

The explained variance ratio for the input and output PCA models is shown in \cref{fig:explained_var_ratio}.

\begin{figure}[h]
    \centering
    \caption{Explained variance ratio for the PCA models trained on the pre-activations (input) and activations (output) of the MLP layers.}
    \label{fig:explained_var_ratio}
\end{figure}

\paragraph{Training the PCA models}
PCA models were trained using Scikit-learn \citep{scikit-learn} on the training split. Input data were centered but not whitened (PCA components were not scaled to unit variance).

\paragraph{Training the symbolic models}
After fitting PCA, we define a Python callable mapping the MLP's reduced-dimensionality pre-activations to its reduced-dimensionality activations, and wrap it with SymTorch for SR. We used a random subset of 6000 samples from the training set. SR parameters follow the defaults in \cref{tab:def_sr_params} with the number of iterations increased to 5000.

\paragraph{Experimental conditions}
We evaluate four conditions, all applied using forward hooks and pre-forward hooks to avoid modifying the surrounding architecture:
\begin{enumerate}
    \item \textbf{Baseline}: no intervention.
    \item \textbf{Skip}: the MLP is replaced with an identity function, effectively removing the layer.
    \item \textbf{PCA}: PCA compression is applied at input and output, but the full-dimensionality MLP is still executed. This isolates the cost of dimensionality reduction alone.
    \item \textbf{SR}: PCA compression is applied and the MLP is replaced with symbolic surrogates.
\end{enumerate}
Layers are selected greedily in order of increasing per-layer perplexity impact (see \cref{tab:per_layer_ppl}).

\paragraph{Efficiency results}

\begin{table}[t]
  \centering
  \caption{Perplexity degradation (on the test set), throughput gain, and memory saved as a
           function of layers skipped, for Skip, SR, and PCA. Memory savings are
           identical across Skip and SR; PCA yields no memory saving.}
  \label{tab:llm-distill-summary}
  \small
  \setlength{\tabcolsep}{5pt}
  \begin{tabular}{
    c
    S[table-format=2.3]
    S[table-format=2.3]
    S[table-format=2.3]
    S[table-format=2.1]
    S[table-format=2.1]
    S[table-format=4.0]
    S[table-format=2.1]
  }
    \toprule
    & \multicolumn{3}{c}{$\Delta$ PPL $\downarrow$}
    & \multicolumn{2}{c}{$\Delta$ Throughput (\%) $\uparrow$}
    & \multicolumn{2}{c}{Memory Saved $\uparrow$} \\
    \cmidrule(lr){2-4} \cmidrule(lr){5-6} \cmidrule(lr){7-8}
    {Layers} & {SR} & {PCA} & {Skip} & {SR} & {Skip} & {(MB)} & {(\%)} \\
    \midrule
    1  &  0.444 &  0.451 &  0.565 &  2.4 &  2.4 &   158 &  2.7 \\
    2  &  0.973 &  1.004 &  1.289 &  4.8 &  4.9 &   315 &  5.4 \\
    3  &  1.651 &  1.653 &  2.249 &  7.4 &  7.5 &   473 &  8.0 \\
    4  &  2.547 &  2.955 &  3.815 &  9.8 & 10.3 &   630 & 10.7 \\
    5  &  3.415 &  4.274 &  4.555 & 12.6 & 13.2 &   788 & 13.4 \\
    6  &  4.256 &  5.637 &  5.669 & 15.4 & 16.2 &   945 & 16.1 \\
    7  &  6.375 & 10.059 &  9.262 & 18.7 & 19.6 &  1103 & 18.7 \\
    8  &  8.463 & 14.050 & 13.132 & 21.8 & 23.0 &  1260 & 21.4 \\
    9  & 13.652 & 36.309 & 26.302 & 25.0 & 26.6 &  1418 & 24.1 \\
    10 & 19.723 & 58.727 & 40.816 & 28.8 & 30.5 &  1575 & 26.8 \\
    \bottomrule
  \end{tabular}
\end{table}

\cref{tab:llm-distill-summary} shows the full results. SR achieves consistently lower perplexity degradation than Skip at equivalent layer counts, demonstrating that the symbolic surrogate genuinely captures MLP behavior. The near-identical perplexity costs of SR and PCA across most operating points confirm that dimensionality reduction --- not symbolic approximation --- is the dominant source of quality loss. Memory savings are identical between Skip and SR, as both methods remove the MLP weight matrices from GPU memory.

\paragraph{Inference benchmarking}
After 5 warm-up passes, we recorded the time for 100 forward passes of the test set through the model with KV caching disabled. Throughput is the average tokens per second across these passes, run on an Nvidia A100-SXM4-80GB GPU. \cref{tab:llm_pca_benchmark} compares the symbolic hybrid against popular open-source LLMs of comparable scale. Even after the perplexity increase from symbolic replacement, the modified Qwen2.5-1.5B model remains on the Pareto front of the throughput--perplexity tradeoff.

\begin{table}[t]
  \caption{Perplexity on the test set and inference benchmarking for various open-source language models and SR-modified Qwen2.5-1.5B variants. SR rows list the greedily selected replaced layers.}
  \label{tab:llm_pca_benchmark}
  \centering
  \resizebox{\textwidth}{!}{%
    \begin{tabular}{lcccc}
      \toprule
      Model & \shortstack{Perplexity on\\ Validation Set} &\shortstack{ Avg. Latency \\ (ms)} & \shortstack{p95 Latency\\ (ms)} & \shortstack{Throughput \\(tokens/s)}  \\
      \midrule
      \rowcolor{gray!20} Qwen2.5-1.5B (Baseline) & 10.79 & 216.99 & 217.13 & 4719.07 \\
      \rowcolor{gray!20} Qwen2.5-1.5B SR (15) & 11.24 & 211.87 & 212.07 & 4833.25 \\
      \rowcolor{gray!20} Qwen2.5-1.5B SR (15, 17) & 11.77 & 207.14 & 207.58 & 4943.49 \\
      \rowcolor{gray!20} Qwen2.5-1.5B SR (15, 17, 10) & 12.44 & 202.16 & 202.41 & 5065.23 \\
      \rowcolor{gray!20} Qwen2.5-1.5B SR (15, 17, 10, 13) & 13.34 & 197.64 & 197.86 & 5181.18 \\
      \rowcolor{gray!20} Qwen2.5-1.5B SR (15, 17, 10, 13, 14) & 14.21 & 193.12 & 193.32 & 5302.48 \\
      \rowcolor{gray!20} Qwen2.5-1.5B SR (15, 17, 10, 13, 14, 16) & 15.05 & 188.16 & 188.57 & 5442.06 \\
      \rowcolor{gray!20} Qwen2.5-1.5B SR (15, 17, 10, 13, 14, 16, 12) & 17.17 & 183.58 & 183.79 & 5578.03 \\
      \rowcolor{gray!20} Qwen2.5-1.5B SR (15, 17, 10, 13, 14, 16, 12, 9) & 19.26 & 179.12 & 179.48 & 5716.79 \\
      \rowcolor{gray!20} Qwen2.5-1.5B SR (15, 17, 10, 13, 14, 16, 12, 9, 11) & 24.44 & 174.42 & 174.79 & 5870.78 \\
      \rowcolor{gray!20} Qwen2.5-1.5B SR (15, 17, 10, 13, 14, 16, 12, 9, 11, 18) & 30.52 & 169.16 & 169.45 & 6053.37 \\
      Qwen2.5-3B-Instruct & 9.91 & 397.85 & 401.55 & 2573.84 \\
      Qwen2.5-7B-Instruct & 8.87 & 866.06 & 867.71 & 1182.36 \\
      Llama-3.1-8B-Instruct & 8.52 & 891.84 & 892.89 & 1148.18 \\
      Llama-3.2-1B-Instruct & 14.21 & 169.31 & 170.08 & 6048.19 \\
      Llama-3.2-3B-Instruct & 12.32 & 414.51 & 416.03 & 2470.36 \\
      SmolLM2-1.7B-Instruct & 10.33 & 240.48 & 240.89 & 4258.23 \\
      SmolLM3-3B & 10.20 & 392.89 & 393.52 & 2606.35 \\
      TinyLlama-1.1B-Chat-v1.0 & 14.80 & 149.00 & 149.44 & 6872.68 \\
      \bottomrule
    \end{tabular}
  }
\end{table}

\paragraph{Limitations}
The following limitations apply to the current experimental setup:
\begin{itemize}
    \item \textbf{Non-additive gains} The symbolic models are fit independently per layer and optimized in isolation. Consequently, gains do not compound additively across layers and greedy selection does not yield a globally optimal solution.
    \item \textbf{Dimensionality reduction bottleneck} The primary source of perplexity degradation is PCA compression rather than the symbolic approximation itself. Further progress depends on reducing the PCA bottleneck without sacrificing quality.
    \item \textbf{Fixed operator set} SR searches over a pre-specified set of mathematical operators. Operators outside this set cannot be discovered, potentially preventing the most compact representation of a given layer's function.
\end{itemize}

\paragraph{Extensions}
Several natural directions extend this work:
\begin{itemize}
    \item \textbf{Post-replacement fine-tuning} Running LoRA fine-tuning after symbolic replacement could allow remaining network weights to adapt globally, potentially recovering some of the perplexity cost.
    \item \textbf{Joint PCA--SR optimization} Rather than fitting PCA and SR sequentially, a joint optimization could discover a lower-dimensional subspace better suited to symbolic approximation.
    \item \textbf{Beyond MLP layers} The same pipeline could in principle be applied to attention projections or other transformer sub-components.
    \item \textbf{Composability with quantization} SR replacement and quantization reduce different kinds of redundancy; combining them may yield compounding efficiency gains.
\end{itemize}



\end{document}